\documentclass[10pt,twocolumn,letterpaper]{article}

\usepackage[pagenumbers]{wacv} % To force page numbers, e.g. for an arXiv version

\usepackage{graphicx}
\usepackage{amsmath}
\usepackage{amssymb}
\usepackage{booktabs}
\usepackage{multirow}
\usepackage{subcaption}
\usepackage[font=small]{caption}
\usepackage[accsupp]{axessibility}

\usepackage[pagebackref,breaklinks,colorlinks]{hyperref}

\usepackage[capitalize]{cleveref}
\crefname{section}{Sec.}{Secs.}
\Crefname{section}{Section}{Sections}
\Crefname{table}{Table}{Tables}
\crefname{table}{Tab.}{Tabs.}

\def\wacvPaperID{384} % *** Enter the WACV Paper ID here
\def\confName{WACV}
\def\confYear{2024}

\begin{document}

%%%%%%%%% TITLE - PLEASE UPDATE
\title{Scene Text Image Super-resolution based on Text-conditional Diffusion Models}

\author{Chihiro Noguchi \qquad\ Shun Fukuda \qquad\ Masao Yamanaka\\
Toyota Motor Corporation, Japan\\
{\tt\small \{chihiro\_noguchi\_aa, shun\_fukuda, masao\_yamanaka\}@mail.toyota.co.jp}
% For a paper whose authors are all at the same institution,
% omit the following lines up until the closing ``}''.
% Additional authors and addresses can be added with ``\and'',
% just like the second author.
% To save space, use either the email address or home page, not both
}
\maketitle

%%%%%%%%% ABSTRACT
\begin{abstract}
Scene Text Image Super-resolution (STISR) has recently achieved great success as a preprocessing method for scene text recognition. STISR aims to transform blurred and noisy low-resolution (LR) text images in real-world settings into clear high-resolution (HR) text images suitable for scene text recognition. In this study, we leverage text-conditional diffusion models (DMs), known for their impressive text-to-image synthesis capabilities, for STISR tasks. Our experimental results revealed that text-conditional DMs notably surpass existing STISR methods. Especially when texts from LR text images are given as input, the text-conditional DMs are able to produce superior quality super-resolution text images. Utilizing this capability, we propose a novel framework for synthesizing LR-HR paired text image datasets. This framework consists of three specialized text-conditional DMs, each dedicated to text image synthesis, super-resolution, and image degradation. These three modules are vital for synthesizing distinct LR and HR paired images, which are more suitable for training STISR methods. Our experiments confirmed that these synthesized image pairs significantly enhance the performance of STISR methods in the TextZoom evaluation. The code is available at \url{https://github.com/ToyotaInfoTech/stisr-tcdm}.
\end{abstract}

\begin{figure}[h!]
    \begin{tabular}{c}
      \begin{minipage}[t]{0.95\linewidth}
        \centering
        \includegraphics[width=8.2cm]{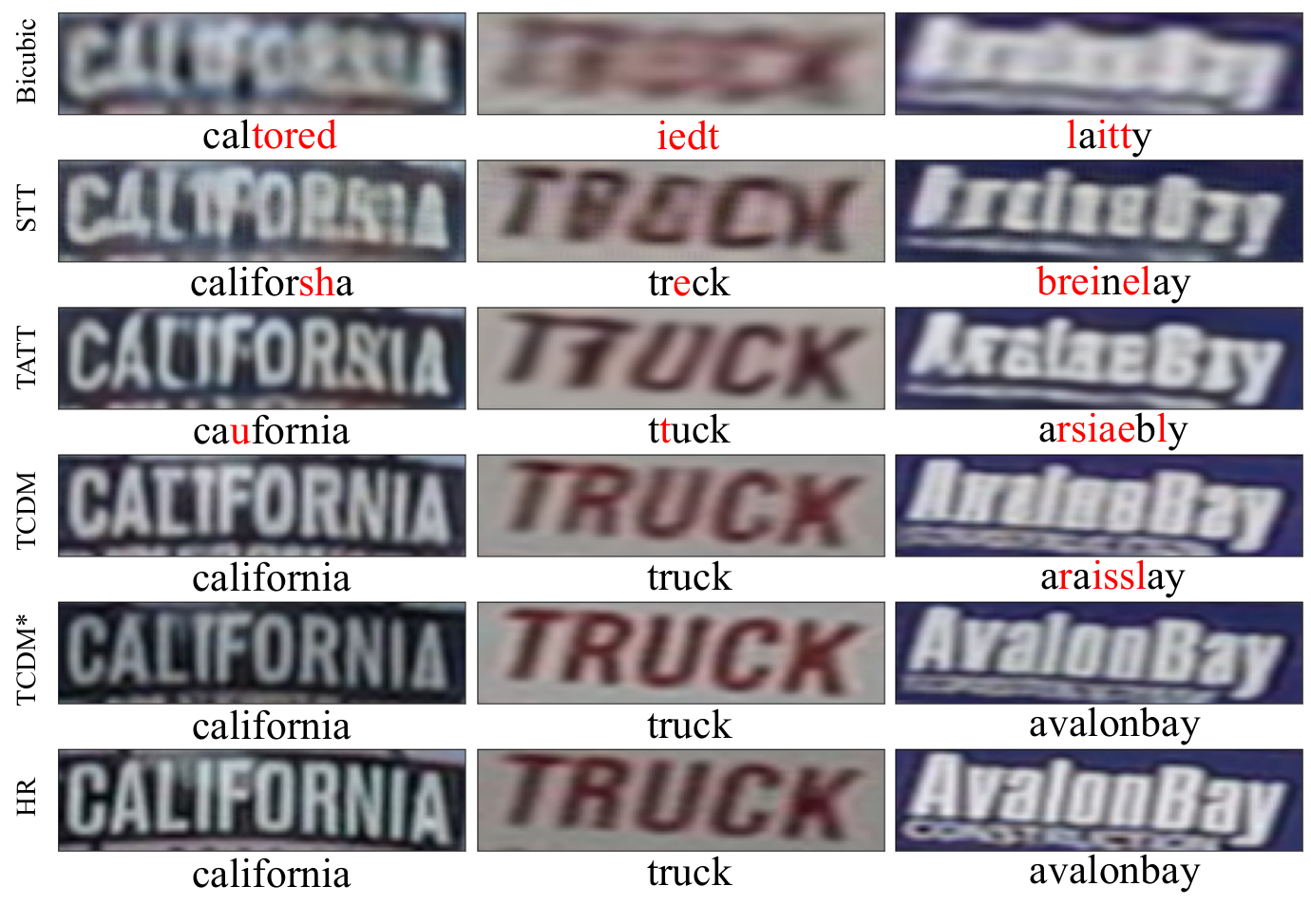}
      \end{minipage}
      
    \end{tabular}
    \caption{SR text images restored from different methods. Text below the images indicates text recognition results. Red characters indicate incorrect or missing results. TCDM denotes the text-conditional diffusion model, while TCDM* indicates TCDM trained using ground-truth text input.}
\label{fig:top_figure}
\end{figure}

%%%%%%%%% BODY TEXT
\section{Introduction}
Scene text recognition has attracted considerable attention because of its high applicability in many areas.
However, it remains a challenging task when the input images are significantly blurred and low-resolution.
To address these problems, scene text image super-resolution (STISR) is a promising approach.
STISR aims to restore the high-resolution (HR) text images from low-resolution (LR) text images.
Unlike standard single-image super-resolution \cite{dong2015image,7780551,Ledig_2017_CVPR,lim2017enhanced,Zhang_2018_CVPR,lai2017deep}, which is applied to general scene images, STISR is specialized for text images. 
Hence, the essence of STISR lies in restoring HR text images while preserving the textual information contained in the LR images.
The restored images are called super-resolution (SR) images.

To effectively train STISR models, there is a need for a substantial quantity of paired LR-HR text images. However, obtaining such paired images from real-world scenarios is significantly expensive, which results in a scarcity of public datasets tailored for STISR. An alternative strategy is the generation of synthetic LR images. This can be achieved by applying synthetic degradation, such as bicubic interpolation and blur kernels, to text images. However, the synthetic images produced via these simple degradation methods might not be apt for real-world settings, where more complex degradations can occur \cite{10.1007/978-3-030-58607-2_38}.

In this study, we aim to produce realistic paired text images that are better suited for training STISR methods. The proposed framework in this study hinges on two primary elements. Firstly, it utilizes text-conditional diffusion models. Diffusion models (DMs) \cite{pmlr-v37-sohl-dickstein15,song2019generative,ho2020denoising} have demonstrated remarkable generative performance in various tasks, especially in text-to-image synthesis \cite{nichol2021glide,ramesh2022hierarchical,saharia2022photorealistic}. This generative prowess can also be considered effective for STISR. Unlike text-to-image synthesis settings, in STISR, text prompts are not predetermined. Therefore, a pretrained text recognition model is required to extract text features from the text images. Our experimental results indicate that while the vanilla DM is sufficiently effective, incorporating text feature extraction further elevates its performance. The second element arises from the observation that the performance of STISR methods can be substantially boosted by using ground-truth texts of LR text images, rather than depending on outputs of the text recognition model.
These two elements allow our framework to generate high-quality super-resolution text images from the provided text images and their corresponding ground-truth texts.

Furthermore, our proposed method integrates two additional components. The first is an image degradation model that generates LR text images from the provided text images. Combined with the aforementioned super-resolution model, our framework can generate LR-HR paired text images from the input text images with their ground-truth texts. The second component is a text image generation model that generates text images from text strings. Once these models are trained, they allows for the generation of as many LR-HR paired images as desired. In our proposed framework, all the three components are implemented using text-conditional DMs.

Synthesising LR-HR paired images is not novel strategy to enhance single image super-resolution performance. BSRGAN \cite{zhang2021designing} and Real-ESRGAN \cite{wang2021real} combine different types of synthetic degradations to emulate realistic ones. Sidiya et al. \cite{10.3389/frsip.2023.1106465} introduce two Style-GAN-based models \cite{karras2020analyzing} for synthesizing paired image. MCinCGAN \cite{zhang2019multiple} and Pseudo-SR \cite{maeda2020unpaired}, which are based on Cycle-GAN \cite{zhu2017unpaired}, eliminate the need for image-level paired training images. However, these methods are not specialized for text images. Conversely, our proposed framework is designed specifically for text images, enabling the generation of high-quality paired images by utilizing the textual content in the text images as input.

The super-resolution and image degradation models in our framework are trained using TextZoom \cite{10.1007/978-3-030-58607-2_38}, the most commonly dataset for STISR. For the text image generation model, we explore two scenarios: one where the model is trained exclusively on TextZoom and another where it is trained on both TextZoom and additional text image datasets. The paired images synthesized by our framework are combined with TextZoom to form augmented datasets. Our experiments reveal that the augmented datasets enhance the performance of STISR methods, showing a marked improvement over relying solely on TextZoom.

Our contributions can be summarized as follows. (1) To the best of our knowledge, this study represents the pioneering effort applying text-conditional DMs to scene text image super-resolution. Experimental validation confirms that the DMs attain state-of-the-art performance when assessed on TextZoom \cite{10.1007/978-3-030-58607-2_38}. (2) We introduce a framework to generate LR-HR paired text images. Using ground-truth texts of input text images for training each component of the framework, it can generate high-quality paired images that are apt for the training datasets of STISR methods. (3) Through our experiments, we show that the augmented datasets effectively enhance the performance of STISR methods, surpassing results obtained by relying solely on TextZoom.

\section{Related Works}

\subsection{Scene Text Recognition}
Scene text recognition methods basically accept a cropped image that contains only a single word as input.
The text images are fed into the CNN- or Transformer-based feature extractor.
Several decoding methods have been proposed using image features to predict characters.
Two major decoding methods exist: CTC decoding \cite{10.1145/1143844.1143891,7801919} and attention decoding \cite{LUO2019109,8395027,Fang2021ReadLH,10.1007/978-3-031-19815-1_11}.
Please refer to the comprehensive survey \cite{10.1145/1143844.1143891} for more details on scene text recognition.

\subsection{Scene Text Image Super-Resolution}
The goal of STISR is to restore an HR image from an LR image.
Unlike standard single image super-resolution, in STISR, one word always appears in the input image (called a text image).
Therefore, extracting accurate textual contents from text images and effectively utilizing it is essential to generate high-quality SR images.
In fact, not only PSNR/SSIM metrics, which are commonly used in single image super-resolution, but also text recognition accuracy are used to evaluate the performance of STISR methods.

Many methods have been proposed in this area. Dong et al. \cite{dong2015boosting} applied an SRCNN \cite{dong2015image} to obtain SR images and demonstrated its effectiveness in terms of scene text recognition performance.
TextSR \cite{DBLP:journals/corr/abs-1909-07113} and MCGAN \cite{8237298} are based on SRGAN \cite{Ledig_2017_CVPR} and use loss functions to adopt the guidance of the text recognizer.
PCAN \cite{10.1145/3474085.3475469} and \cite{9040515} focused on the high-frequency components of an image during reconstruction.
PlugNet \cite{10.1007/978-3-030-58555-6_10} and TSRN \cite{10.1007/978-3-030-58607-2_38} utilized novel modules based on residual blocks to enhance SR quality.
In STT \cite{Chen_2021_CVPR}, position- and content-aware losses were proposed.
These losses encourage the model to focus on character regions and labels, which helps produce SR text images suitable for text recognition.
TATT \cite{ma2021text,Ma_2022_CVPR} incorporates text prior information obtained from a text recognizer into a transformer-based network using cross-attention modules.
C3-STISR \cite{zhao2022c3} exploits three clues: text prior, visual, and linguistical information.
TG \cite{chen2022text} was proposed to concentrate on the stroke-level internal structures of characters.
Wang et al. \cite{10.1007/978-3-030-58607-2_38} proposed TextZoom, which contains real LR-HR paired images and is widely used to evaluate the performance of STISR methods.

\subsection{Diffusion Models}
Gaussian DMs were first introduced in \cite{pmlr-v37-sohl-dickstein15} and have been improved for image generation \cite{ho2020denoising,NEURIPS2021_49ad23d1,pmlr-v139-nichol21a,ho2022cascaded} and various other tasks \cite{10.1145/3528223.3530104,baranchuk2022labelefficient}.
DMs achieve significant performance, particularly in text-to-image synthesis.
A major concern here is how to condition the DM to control the image generated using text prompts accurately.
To this end, classifier guidance was proposed in \cite{NEURIPS2021_49ad23d1}, which can improve the sample quality using class labels while reducing the diversity of the generated images.
Classifier guidance was extended to classifier-free guidance in \cite{ho2021classifierfree}, eliminating the need for an additional classifier.
This makes it easier to condition DMs on information that is difficult to handle with a classifier such as text \cite{nichol2021glide,ramesh2022hierarchical,saharia2022photorealistic,ramesh2022hierarchical}.
In addition, DMs have been applied to single image super-resolution \cite{9887996}.
The LR image is created by downsampling the HR image, concatenating it with the input image, and feeding it to the model.
Additionally, SR is often used as an auxiliary task in text-to-image synthesis to generate high-resolution images \cite{nichol2021glide}.

\section{Methodology}
The method proposed in this paper is based on denoising diffusion probabilistic models (DDPMs).
In the following, we describe the improvement of DDPMs for application to STISR in Sec. \ref{sec:conditional_ddpm}.
Loss functions for extracting textual information and the model architecture of the proposed method are described in Sec. \ref{sec:tp_generator_training} and \ref{sec:model_arcitecture}, respectively.
Next, we introduce a framework for synthesizing LR-HR paired text images in Sec. \ref{sec:synthetic_lr_hr_pair_img_generation}.

\subsection{Text-Conditional DMs for STISR}
\label{sec:conditional_ddpm}
A DM consists of forward and reverse processes.
In the forward process, Gaussian noise is gradually added to the input image, and eventually it becomes pure Gaussian noise.
Conversely, in the reverse process, starting from pure Gaussian noise, noise is sequentially removed to recreate the original image.

The objective here is to achieve each step of the reverse process from a Gaussian noise input.
Recent successful models view this problem as one of predicting the Gaussian noise contained at each step of the forward process with a sequence of denoising autoencoders
$\epsilon_\theta(x_t,t)$, $\forall t\in\{1,\dots,T\}$.
Here, $x_t$ is the output of the $t$th step of the forward process.
Consequently, the loss function is given as 
\begin{equation}
    \mathcal{L}_{u}=\mathbb{E}_{t,x_0,\epsilon}[\|\epsilon-\epsilon_\theta(x_t,t)\|^2],
    \label{eq:unconditional_ddpm_loss}
\end{equation}
where $t$ is sampled uniformly from $\{1,\dots,T\}$ and $\epsilon\sim\mathcal{N}(0,\boldsymbol{I})$.
Note that $x_T$ corresponds to a noise image.
The derivation of Eq. \ref{eq:unconditional_ddpm_loss} is provided in Sec. \ref{sec:ddpm} and \cite{ho2020denoising}.

From the STISR perspective, it is necessary to control the output HR text image $x_0$ under two different conditions: LR text image and text in the text image.
According to prior studies on text-to-image synthesis, this can be easily achieved by simply conditioning $\epsilon_\theta(x_t,t)$ in Eq. \ref{eq:unconditional_ddpm_loss}.

When conditioning a DM on an LR image $x^l$, $x_t$ is replaced with $x^\prime_t=[x_t,x^l]$, where $[\cdot,\cdot]$ indicates the operation of concatenation in the channel dimension. Here, $x^l$ must be bicubic interpolated to the size of $x_t$.

Unlike text-to-image synthesis, text prompts are not provided by users in STISR.
Therefore, textual information must be extracted from the provided text images.
In previous studies \cite{ma2021text,Ma_2022_CVPR}, such textual information, which is referred to as {\it text prior}, is extracted using a pretrained text recognition method.
This model is called {\it text prior generator}.
In this study, we additionally introduce ground-truth text prior, which is created from user-provided text instead of using the text prior generator.
The text prior or the ground-truth text prior is then fed into a text encoder to obtain text features $z$.
Consequently, the loss function in Eq. \ref{eq:unconditional_ddpm_loss} is modified as follows:

\begin{equation}
    \mathcal{L}_{c}=\mathbb{E}_{t,x_0,\epsilon}[\|\epsilon-\epsilon_\theta(x^\prime_t,z,t)\|^2].
    \label{eq:conditional_ddpm_loss}
\end{equation}

\begin{figure*}[h!]
    \begin{tabular}{c}
      \begin{minipage}[t]{0.95\linewidth}
        \centering
        \includegraphics[width=16.0cm]{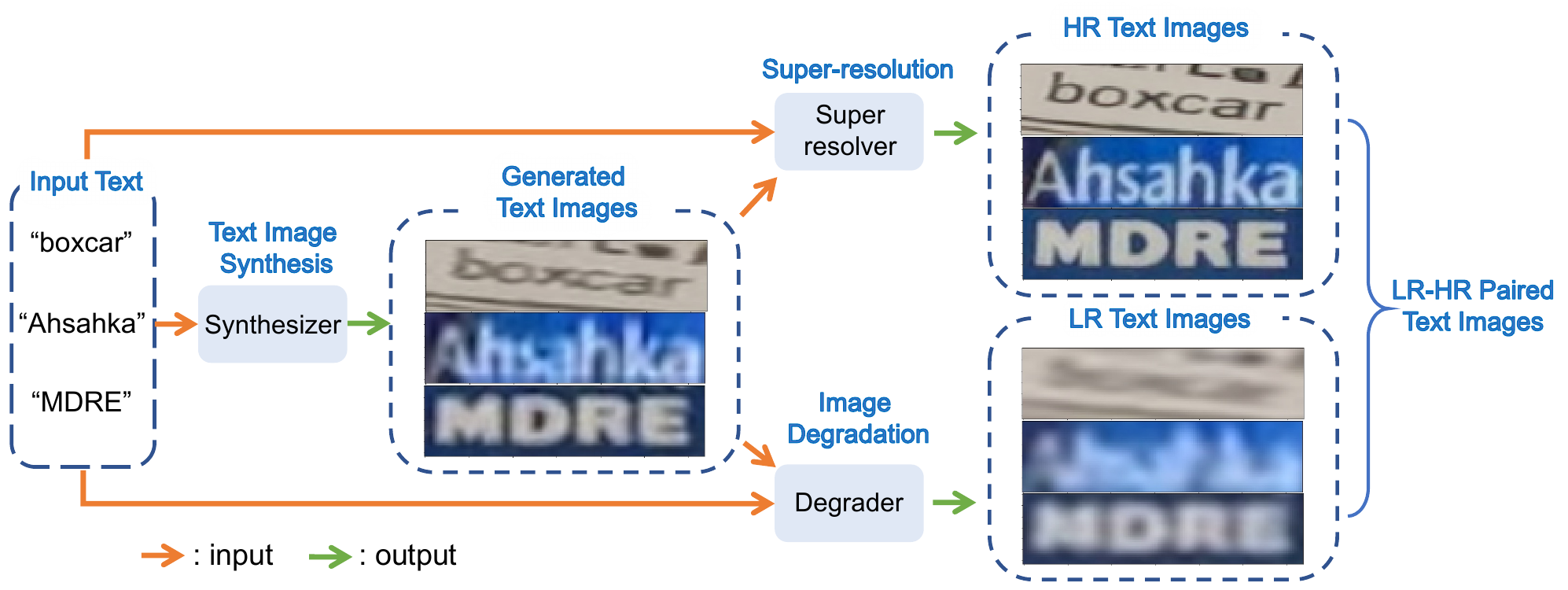}
      \end{minipage}
    \end{tabular}
    \caption{Overview of the proposed framework for LR-HR paired text image synthesis. First, text images are generated by Synthesizer with user-provided text. Next, the generated text images are fed to Super-resolver and Degrader to generate HR and LR images, respectively. Synthesizer, Super-resolver, and Degrader are based on text-conditional DMs, which take user-provided texts as input.}
\label{fig:overview}
\end{figure*}

\subsubsection{Text Prior Generator}
\label{sec:tp_generator_training}
Following prior studies \cite{ma2021text,Ma_2022_CVPR,zhao2022c3}, the text prior generator is trained concurrently with the DM.
We use intermediate features from one layer before the final output as text features.
As the input to the text prior generator, we can use $t$th output image $x_t$ in addition to LR image $x^l$.
When $t$ is small, $x_t$ can be sufficiently clear to be useful for the text prior generator, whereas when $t$ is large, $x_t$ may still be noisy.
Therefore, we consider the weighted sum of the two images $y_t=(1-(t/T)^3)x_t + (t/T)^3x^l$ as the input for the text prior generator.
When $t$ is large, $y_t$ is close to $x^l$; as $t$ decreases, $y_t$ approaches $x_t$.

When ground-truth text prior is used, a text prior generator is not required.
Instead, the ground-truth text prior must be manually created from the corresponding texts and is represented by a matrix $f_P\in\mathbb{R}^{l\times |\mathcal{A}|}$, where $l$ is the maximum length of the input text, and $\mathcal{A}$ is the character set.
Each row of $f_P$ is a one-hot vector.
When the text length is less than $l$, all zero vectors are inserted at equal intervals to fit the size of $f_P$.

\subsubsection{Model Architecture}
\label{sec:model_arcitecture}
In our text-conditional DMs, we adopt the prevailing UNet architecture \cite{10.1007/978-3-319-24574-4_28} similar to previous studies \cite{pmlr-v139-nichol21a,NEURIPS2021_49ad23d1}. However, unlike text-image synthesis, STISR does not require a large language model for the text encoder because the input text is typically only a single word. Therefore, we adopt a simple self-attention-based architecture for the text encoder (see Sec. \ref{sec:design_of_dimss_arc} for more details).

To condition DMs on text features obtained from the text encoder, cross-attention modules \cite{NIPS2017_3f5ee243} have been a common choice in text-to-image synthesis. In previous studies, cross-attention modules are inserted in multiple resolution layers \cite{saharia2022photorealistic,Rombach_2022_CVPR}. On the other hand, in our text-conditional DMs, cross-attention modules are inserted only at the bottom of the UNet. Our experimental results show that this architecture can sufficiently grasp the text features while curbing computational expenses. However, a single cross-attention module is inadequate, thus we adopt an architecture with multiple cross-attention modules. See Sec. \ref{sec:design_of_dimss_arc} for the architecture details.

\begin{table*}[!t]
  \centering
  \resizebox{0.95\textwidth}{!}{
  \begin{tabular}{c||cccc|cccc|cccc}
    \hline
    \multirow{2}{*}{Method} & \multicolumn{4}{c|}{CRNN \cite{7801919}} & \multicolumn{4}{c|}{MORAN \cite{LUO2019109}} &  \multicolumn{4}{c}{ASTER \cite{8395027}} \\
    \cline{2-13}
    & Easy & Medium & Hard & Avg. & Easy & Medium & Hard & Avg. & Easy & Medium & Hard & Avg. \\
    \hline
  Bicubic & 36.4\% & 21.1\% & 21.1\% & 26.8\%  & 60.6\% & 37.9\% & 30.8\% & 44.1\% & 67.4\% & 42.4\% & 31.2\% & 48.2\% \\
  HR & 76.4\% & 75.1\% & 64.6\% & 72.4\% & 91.2\% & 85.3\% & 74.2\% & 84.1\% & 94.2\% & 87.7\% & 76.2\% & 86.6\% \\
  \hline
  SRCNN \cite{dong2015image} & 41.1\% & 22.3\% & 22.0\% & 29.2\% & 63.9\% & 40.0\% & 29.4\% & 45.6\% & 70.6\% & 44.0\% & 31.5\% & 50.0\% \\
  SRResNest \cite{Ledig_2017_CVPR} & 45.2\% & 32.6\% & 25.5\% & 35.1\% & 66.0\% & 47.1\% & 33.4\% & 49.9\% & 69.4\% & 50.5\% & 35.7\% & 53.0\% \\
  TSRN \cite{10.1007/978-3-030-58607-2_38} & 52.5\% & 38.2\% & 31.4\% & 41.4\% & 70.1\% & 55.3\% & 37.9\% & 55.4\% & 75.1\% & 56.3\% & 40.1\% & 58.3\% \\
  STT \cite{Chen_2021_CVPR} & 59.6\% & 47.1\% & 35.3\% & 48.1\% & 74.1\% & 57.0\% & 40.8\% & 58.4\% & 75.7\% & 59.9\% & 41.6\% & 60.1\% \\
  PCAN \cite{10.1145/3474085.3475469} & 59.6\% & 45.4\% & 34.8\% & 47.4\% & 73.7\% & 57.6\% & 41.0\% & 58.5\% & 77.5\% & 60.7\% & 43.1\% & 61.5\% \\
  TG \cite{chen2022text} & 61.2\% & 47.6\% & 35.5\% & 48.9\% & 75.8\% & 57.8\% & 41.4\% & 59.4\% & 77.9\% & 60.2\% & 42.4\% & 61.3\% \\
  TATT \cite{Ma_2022_CVPR} & 62.6\% & 53.4\% & 39.8\% & 52.6\% & 72.5\% & 60.2\% & 43.1\% & 59.5\% & 78.9\% & 63.4\% & 45.4\% & 63.6\%  \\
  C3-STISR \cite{zhao2022c3} & 65.2\% & 53.6\% & 39.8\% & 53.7\% & 74.2\% & 61.0\% & 43.2\% & 60.5\% & 79.1\% & 63.3\% & 46.8\% & 64.1\% \\
  \hline
  DDPM* & 66.8\% & 56.5\% & 41.8\% & 55.0\% & \textbf{78.4\%} & 62.2\% & 45.3\% & 62.0\% & 81.1\% & 64.3\% & 48.9\% & 64.7\% \\
  TCDM & \textbf{67.3\%} & \textbf{57.3\%} & \textbf{42.7\%} & \textbf{55.7\%} & 77.6\% & \textbf{62.9\%} & \textbf{45.9\%} & \textbf{62.2\%} & \textbf{81.3\%} & \textbf{65.1\%} & \textbf{50.1}\% & \textbf{65.5\%} \\
    \hline
  \end{tabular}
  }
\caption{Comparison with the existing methods in terms of the recognition accuracy on TextZoom. TCDM denotes the text-conditional DM. * indicates DDPM \cite{pmlr-v139-nichol21a} trained on TextZoom by ourselves.}
\label{table:comparison_with_sota_acc}
\end{table*}

\subsection{LR-HR Paired Text Image Synthesis}
\label{sec:synthetic_lr_hr_pair_img_generation}
In this subsection, we explore the use of text-conditional DMs to build a framework for synthesizing LR-HR paired text images. As shown in Fig.  \ref{fig:overview}, the proposed framework comprises three distinct subtasks: (1) text image synthesis, (2) super-resolution, and (3) image degradation. We refer to the text-conditional DMs for these three tasks as (1) \textit{Synthesizer}, (2) \textit{Super-resolver}, and (3) \textit{Degrader}, respectively. The workflow of the proposed framework is as follows. First, Synthesizer generates text images from user-provided texts. Then, these generated text images are fed to Super-resolver and Degrader to generate HR and LR images, respectively. As a result, from an arbitrary text input, the corresponding LR-HR paired text images can be created. The three text-conditional DMs do not share parameters and depend on different inputs. Synthesizer is solely conditioned on input texts, whereas Super-resolver and Degrader are conditioned both on the text images from Synthesizer and the initial input texts.

Training Synthesizer requires only the text images and their corresponding texts. In other words, it can be trained using scene text recognition datasets, which are more readily available in large volumes than STISR datasets. It is noteworthy that using Synthesizer is optional. Scene text recognition datasets can be directly used as input to Super-resolver and Degrader, instead of using text images generated by Synthesizer. Regarding the training of Super-resolver and Degrader, both paired LR-HR text images and their corresponding texts are required. The Degrader training can be realized by simply swapping the condition with the target image in the Super-resolver training. For the Degrader training, it is not trivial whether to use texts as input. Nonetheless, our experimental results suggest an enhancement in the quality of the generated text images. Refer to Sec. \ref{sec:effectiveness_of_texts_in_degreader} for more details.

Incorporating both Super-resolver and Degrader is crucial to our framework. In scene images, text areas frequently occupy small portions. As a result, text image datasets encompass many low-resolution and blurred images as well as high-resolution ones. By applying these two modules to the provided text images, our framework can generate distinct pairs of LR and HR images.

\begin{table}[!t]
  \centering
  \resizebox{0.45\textwidth}{!}{
  \begin{tabular}{c||c|ccc}
    \hline
    \multirow{2}{*}{Method} & GT & \multirow{2}{*}{SSIM ($\times 10^{-2}$)} & \multirow{2}{*}{PSNR} & \multirow{2}{*}{Acc. (\%)} \\
    & Text & & &  \\
    \hline
  Bicubic &  & 69.61 &  20.35 & 26.8 \\
  HR & & - & - & 72.4 \\
    \hline
  SRCNN \cite{dong2015image} &  & 72.27 & 20.78 & 29.2 \\
  SRResNest \cite{Ledig_2017_CVPR} &  & 74.03 & 21.03 & 35.1 \\
  TSRN \cite{10.1007/978-3-030-58607-2_38} &  & 76.90 & 21.42 & 41.4 \\
  STT \cite{Chen_2021_CVPR} &  & 76.14 & 21.05 & 48.1 \\
  PCAN \cite{10.1145/3474085.3475469} &  & 77.52 & 21.49 & 47.4 \\
  TG \cite{chen2022text} &  & 74.56 & 21.40 & 48.9 \\
  TATT \cite{Ma_2022_CVPR} &  & 79.30 & 21.52 & 52.6 \\
  C3-STISR \cite{zhao2022c3} &  & 77.21 & 21.51 & 53.7 \\
  \hline
  DDPM* &  & 79.50 & 22.70 & 55.0 \\
  TCDM &  & \textbf{79.58} & \textbf{22.83} & \textbf{55.7}\\
    \hline
  TATT & \checkmark & 79.34 & 22.35 & 61.0 \\
  TCDM & \checkmark  & \textbf{80.25} & \textbf{22.86} & \textbf{68.1} \\
  \hline
  \end{tabular}
  }
\caption{Comparison with the existing methods in terms of average of SSIM/PSNR and recognition accuracy on TextZoom. The recognition accuracy was evaluated by CRNN. The bottom two lines show the results of the methods using ground-truth texts.}
\label{table:comparison_with_sota_ssim_psnr}
\end{table}

\section{Evaluation}
\label{sec:experimental_results}

\subsection{Evaluation on TextZoom}
\label{sec:evaluation_on_stisr}

In this subsection, we evaluate the performance of our text-conditional DMs on TextZoom and compare them with state-of-the-art methods in STISR.

\noindent
\textbf{Training Details.}\quad
The size of the HR images was $32\times 128$, and the LR images were bicubic interpolated to the same size before being fed to the DMs.
Following prior studies, we used a character set consisting of digits and lowercase letters plus a blank character when training the text prior generator ($|\mathcal{A}|=37$).
When training with ground-truth text prior, it was created with a character set consisting of digits and lower- and upper-case letters plus a black character ($|\mathcal{A}|=63$).
Additionally, we set the maximum length of the input text $l=26$.
To fine-tune the text prior generator, we used the KL loss \cite{ma2021text}.
Other detailed hyperparameters are provided in Sec. \ref{sec:design_of_dimss_arc}.
All experiments were performed using a workstation equipped with A100 GPUs.

\noindent
\textbf{Dataset.}\quad
To train and evaluate the STISR methods, we used TextZoom \cite{10.1007/978-3-030-58607-2_38}, which contains 21,740 LR-HR pair images and their corresponding text labels.
Images in TextZoom were captured by cameras with different focal lengths in the wild.
The training set of TextZoom consisted of 17,367 pairs, and the rest were used as the test set.
The test set was divided into three subsets according to the camera focal length: easy (1,619 pairs), medium (1,411 pairs), and hard (1,343 pairs).

\noindent
\textbf{Evaluation Metrics.}\quad
We used three metrics to evaluate the performance of STISR methods.
The first two metrics are SSIM and PSNR, which are widely used in single image super-resolution to measure similarity with HR images.
The third one is text recognition accuracy.
Following prior studies, we used three text recognition methods, CRNN \cite{7801919}, MORAN \cite{LUO2019109} and ASTER \cite{8395027} to evaluate the recognition accuracy.

\subsubsection{Comparison with State-of-the-art Methods}
Table \ref{table:comparison_with_sota_acc} shows the comparison results with existing methods in terms of the recognition accuracy on TextZoom.
DDPM corresponds to the vanilla DMs without text conditions.
The two DM-based models outperformed the existing methods, and our text-conditional DM achieved an even better performance than DDPM.
In addition, Tab. \ref{table:comparison_with_sota_ssim_psnr} shows the results in terms of the average SSIM/PSNR and recognition accuracy for the three difficulty levels of the TextZoom test sets.
Notably, the text-conditional DM outperforms the existing methods.
The bottom two rows present the results of TATT and the text-conditional DM when the ground-truth text prior is used.
We can see that their performances are significantly improved by using ground-truth texts.

\begin{figure*}[t!]
    \begin{tabular}{c}
      \begin{minipage}[t]{0.475\linewidth}
        \centering
        \includegraphics[width=7.5cm]{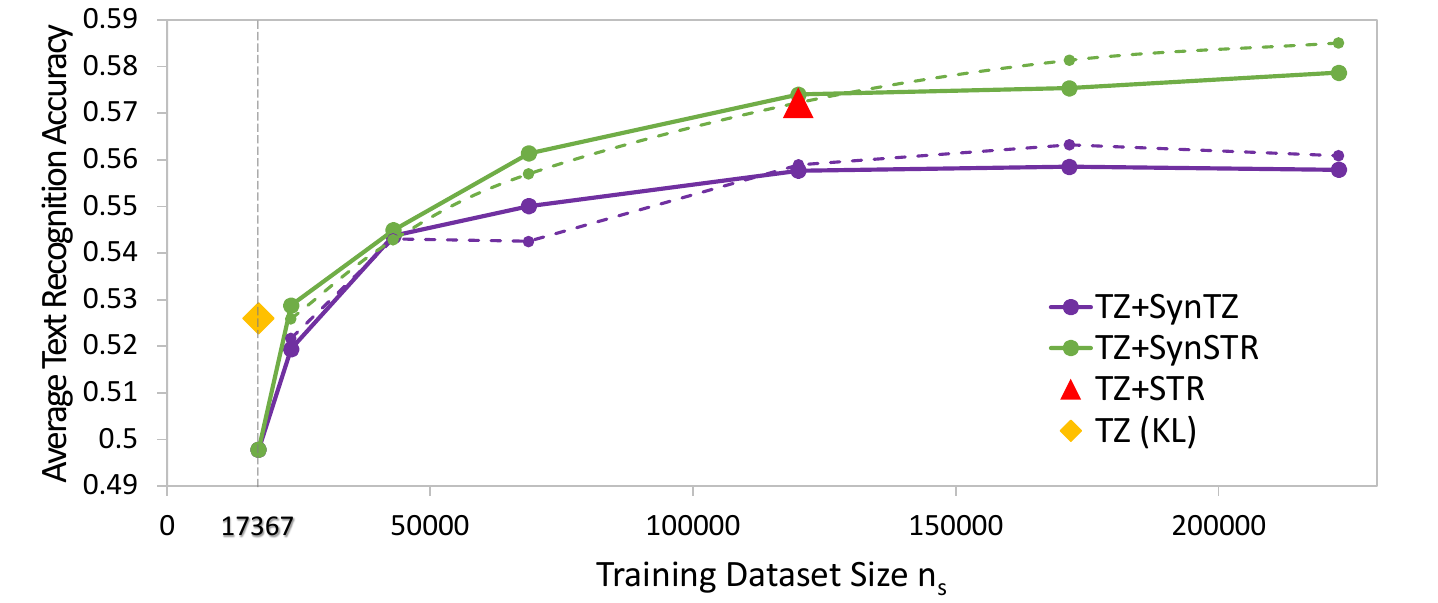}
        \subcaption{}
        \label{fig:acc_dataset_size}
      \end{minipage}

      \begin{minipage}[t]{0.475\linewidth}
        \centering
        \includegraphics[width=7.5cm]{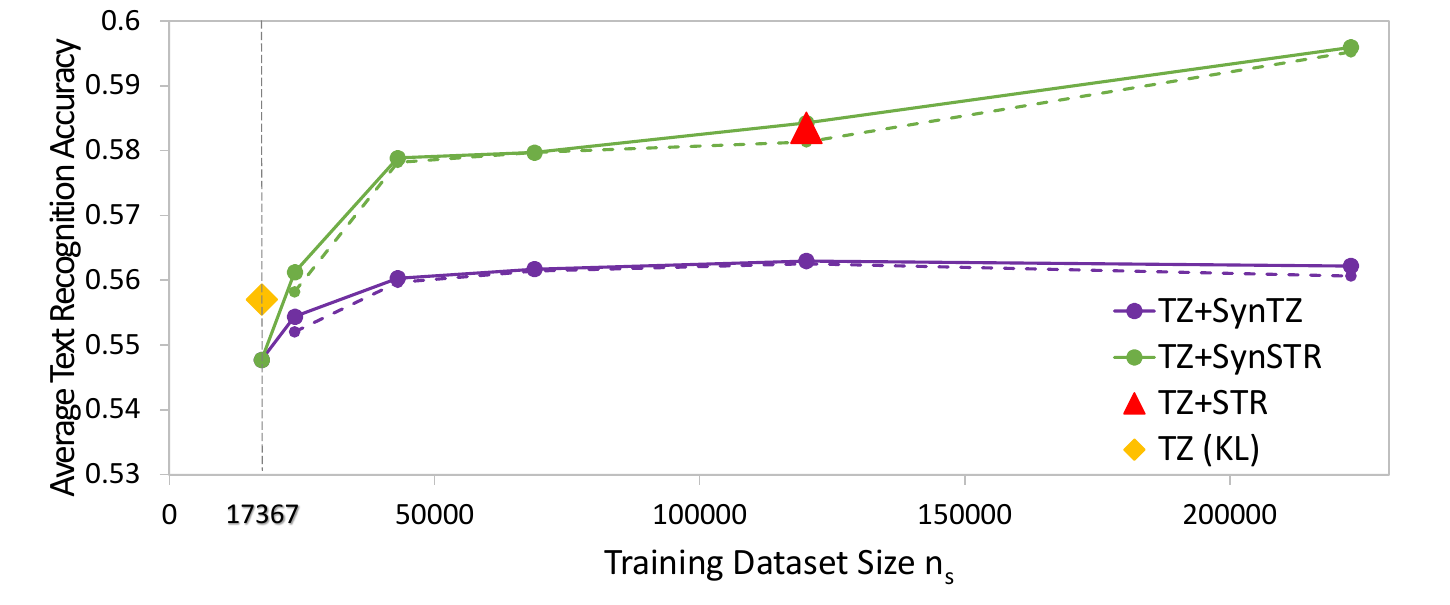}
        \subcaption{}
        \label{fig:acc_dataset_size_dim}
      \end{minipage}
      
        \\
      \begin{minipage}[t]{0.475\linewidth}
        \centering
        \includegraphics[width=7.5cm]{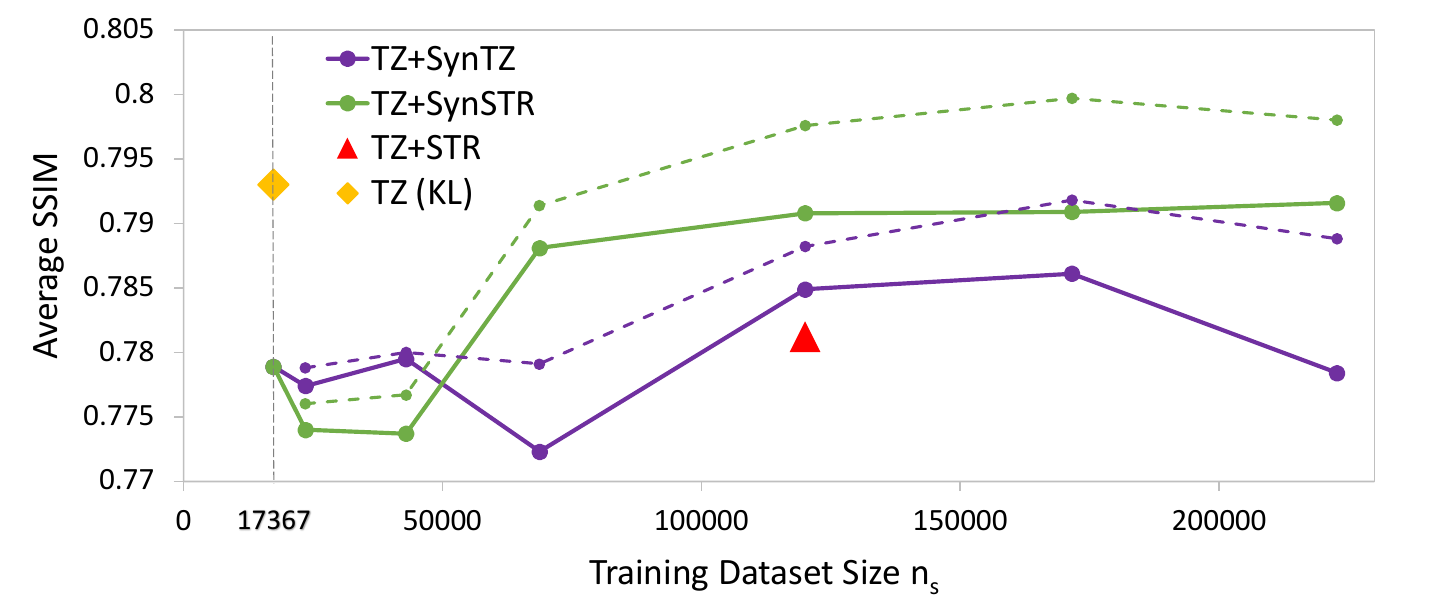}
        \subcaption{}
        \label{fig:ssim_dataset_size}
      \end{minipage}

      \begin{minipage}[t]{0.475\linewidth}
        \centering
        \includegraphics[width=7.5cm]{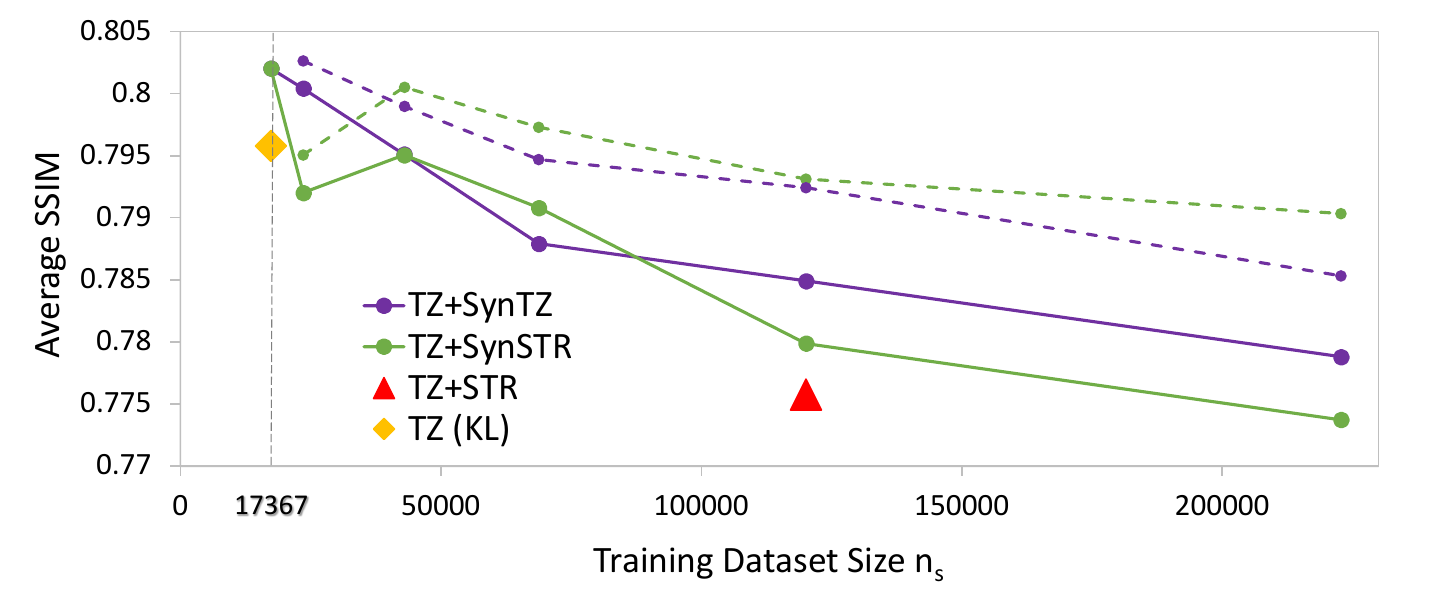}
        \subcaption{}
        \label{fig:ssim_dataset_size_dim}
      \end{minipage}
      
    \end{tabular}
    \caption{Evaluation results of TATT and the text-conditional DM trained on the augmented datasets. (a) and (b) show average recognition accuracy of TATT and the text-conditional DM, respectively. (c) and (d) show average SSIM of TATT and the text-conditional DM, respectively. Dotted lines show the results of fine-tuning using TextZoom only. The solid and dotted lines in the same color correspond to the same augmented dataset. CTC loss was used to train the text prior generator except in the case of ``TZ (KL).'' Here, ``TZ (KL)'' indicates the case where KL loss was used. The size of TextZoom is 17,367.}
\label{fig:acc_ssim_dataset_size}
\end{figure*}

\subsection{Evaluation on Augmented Datasets}
\label{sec:evalution_on_extended_dataset}
We present the experimental results to demonstrate the effectiveness of the LR-HR paired text images synthesized by the proposed framework.
We augmented TextZoom with the synthesized images and trained our text-conditional DM and the state-of-the-art model TATT \cite{Ma_2022_CVPR} on the augmented dataset.
The TextZoom test set was used for the performance evaluation.
Examples of the generated text images are shown in Fig. \ref{fig:lr_hr_examples}.

\noindent
\textbf{Training Details.}\quad
For text inputs to our framework, we randomly chose the length of a word from 2 to 13 and then randomly selected an English word of that length from a dictionary. A comparison based on various maximum word lengths is available in Sec. \ref{sec:word_length_evaluation}. Words comprised solely of digits were included with a 10\% probability. The character set consists of digits as well as lowercase and uppercase letters. To fine-tune the text prior generator, CTC loss \cite{10.1145/1143844.1143891} was used. Hereafter, recognition accuracy is evaluated by CRNN and presented as an average of the results for the three difficulty levels of the TextZoom test set. All other experimental conditions and evaluation metrics were the same as those in Sec. \ref{sec:evaluation_on_stisr}.

\noindent
\textbf{Datasets.}\quad
When training Super-resolver and Degrader, we used LR-HR paired text images of TextZoom.
For training Synthesizer, two datasets were prepared: the first is the HR images of TextZoom, and the second is the HR images of TextZoom plus datasets for scene text recognition.
The scene text recognition datasets consist of 11 real labeled datasets, including SVT \cite{6126402}, IIIT \cite{Mishra2009SceneTR}, IC13 \cite{6628859}, IC15 \cite{7333942}, COCO \cite{veit2016coco}, RCTW \cite{shi2017icdar2017}, Uber \cite{zhang2017uber}, ArT \cite{chng2019icdar2019}, LSVT \cite{sun2019icdar}, MLT19 \cite{nayef2019icdar2019}, and ReCTS \cite{zhang2019icdar}.
We used the above datasets with preprocessing conducted by the authors of \cite{baek2021if} to remove irregular text images (e.g., those containing non-English characters or vertical text).
The preprocessed scene text recognition dataset contains 276K text images.

\begin{figure}[t!]
    \begin{tabular}{c}
      \begin{minipage}[t]{0.95\linewidth}
        \centering
        \includegraphics[width=8.0cm]{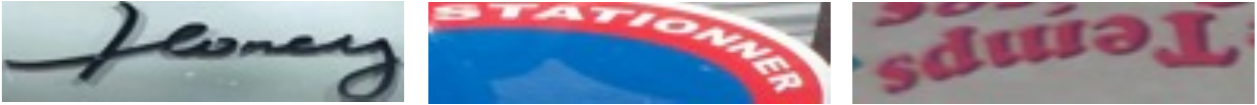}
      \end{minipage}
      
    \end{tabular}
    \caption{Examples of artistic, curved, or rotated text images excluded by preprocessing.}
\label{fig:difficult_samples}
\end{figure}

\noindent
\textbf{Preprocessing for Synthesizer.}\quad
The preprocessed dataset introduced earlier contains some text images that are difficult to handle in the proposed framework, as shown in Fig. \ref{fig:difficult_samples}.
To remove these text images, we used CRNN \cite{7801919} pretrained with synthetic datasets \cite{gupta2016synthetic,mjsynth}, which has poor recognition ability for non-simple text images, as shown in Fig. \ref{fig:difficult_samples}.
We removed text images whose text labels did not match those predicted by CRNN.
As a result, the preprocessed dataset contains 103K text images.

\noindent
\textbf{Postprocessing for Synthesized Text Images.}\quad
In text image synthesis, there were some instances where the input texts did not align with the texts in the synthesized images.
Therefore, we conducted postprocessing similar to preprocessing.
Specifically, ASTER \cite{8395027} was used to predict the texts of the synthesized text images.
If the predicted texts did not match the input texts, the synthesized image was removed.
We believe that ASTER is well-suited for this postprocessing because it can achieve state-of-the-art performance and has no strong language model to correct incorrect spellings.

\subsubsection{Effectiveness of Dataset Augmentation}
\label{sec:effectiveness_of_daatset_extension}
For the preparation of text images, we explored three methods. (1) using the preprocessed scene text recognition dataset introduced earlier. (2) using Synthesizer trained exclusively on TextZoom. (3) using Synthesizer trained on both TextZoom and the preprocessed scene text recognition dataset. We named these three datasets STR, SynTZ, SynSTR, respectively. The resulting text images served as input to Super-resolver and Degrader, generating corresponding paired images.  These paired images were then combined with the original TextZoom (TZ) to form augmented datasets. Consequently, the first augmented dataset is termed TZ+STR, the second as TZ+SynTZ, and the third as TZ+SynSTR.
We denote the size of the augmented dataset as $n_s$ and that of TextZoom as $n_t$.

\textbf{TATT}\quad was first used to evaluate the augmented dataset.
Figure \ref{fig:acc_dataset_size} shows the average recognition accuracy versus $n_s$ as solid lines.
Recognition accuracy tends to improve as $n_s$ increases.
In addition, the recognition accuracy when trained on TZ+SynSTR was higher than when trained on TZ+SynTZ.
TZ+STR achieved an accuracy similar to TZ+SynSTR with the same dataset size.
Additionally, Fig. \ref{fig:ssim_dataset_size} shows SSIM versus $n_s$.
SSIM did not improve depending on $n_s$.
Moreover, SSIM is worse overall compared with the case where $n_s=n_t$ and KL loss was used.
However, this can be improved by fine-tuning with TextZoom only.
Fine-tuning results are shown as dotted lines.
We can see that fine-tuning is effective when the augmented dataset size is large.
The results of PSNR are provided in Sec. \ref{sec:psnr_evaluation}.

\textbf{The text-conditional DM}\quad was next used to evaluate the augmented dataset.
Figures \ref{fig:acc_dataset_size_dim} and \ref{fig:ssim_dataset_size_dim} show the average recognition accuracy and SSIM versus $n_s$, respectively.
Similarly to the results of TATT, recognition accuracy tends to improve as $n_s$ increases.
However, SSIM tends to decrease as $n_s$ increases.
This decrease of SSIM can be attributed the difference that HR images of TextZoom contain many blurred images while those of the augmented dataset do not.
We consider that this is not a negative effect because removing blurred images can improve recognition accuracy.
Additional experimental results on this trade-off are provided in Sec. \ref{sec:higher-resolution_text_images}.
Similarly to the results of TATT, the decrease of SSIM can be reduced by fine-tuning with TextZoom only.

\begin{figure*}[t!]
    \begin{tabular}{c}
      \begin{minipage}[t]{0.95\linewidth}
        \centering
        \includegraphics[width=16.cm]{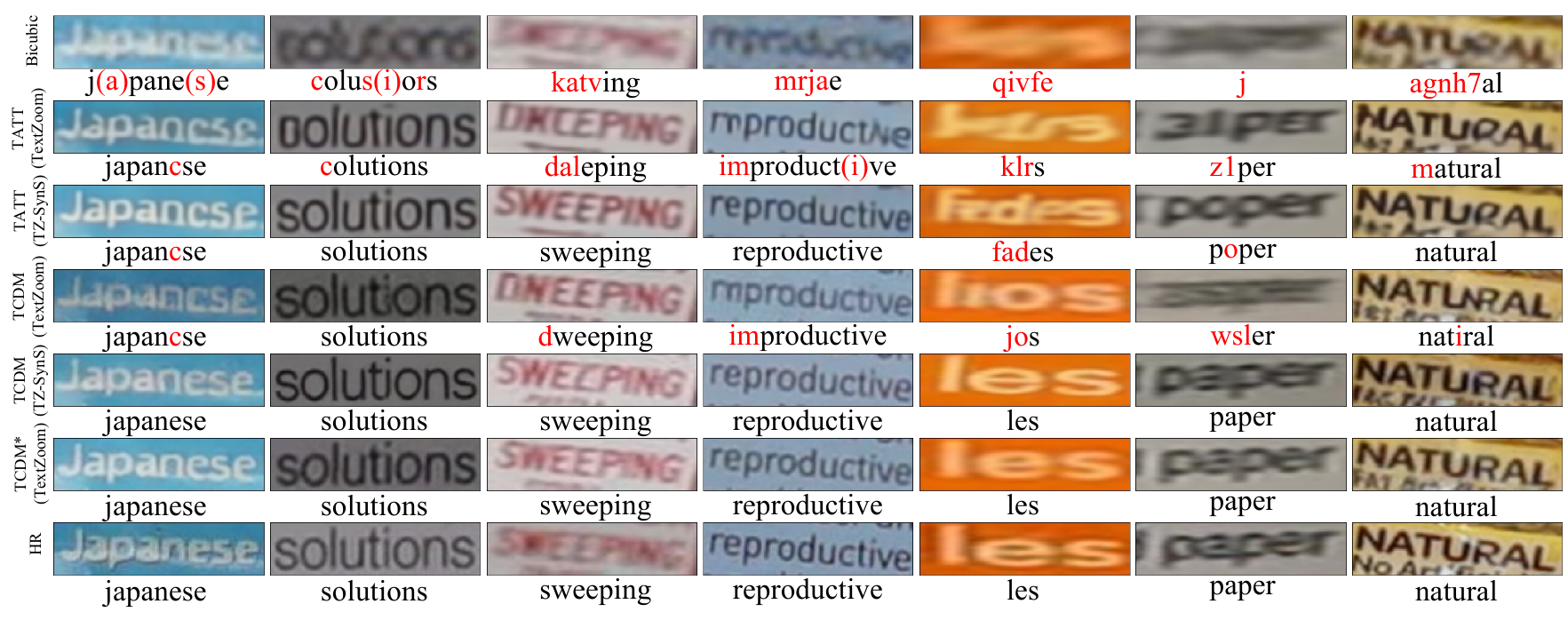}
      \end{minipage}
      
    \end{tabular}
    \caption{Examples of SR text images generated from TATT and text-conditional DMs. The training datasets used are shown in parentheses. Red characters indicate incorrect or missing results.}
\label{fig:lr_hr_examples}
\end{figure*}

\section{Ablation Studies}

\subsection{Comparison to Synthetic Degradation Methods}
Synthetic LR images can be obtained by applying simple degradation techniques, such as bicubic interpolation and blur kernels. In this subsection, we compare the paired text images generated by these simple degradation techniques with those generated by the proposed framework. We employed synthetic degradation pipelines presented in BSRGAN  \cite{zhang2021designing} and Real-ESRGAN \cite{wang2021real} to generate synthetic LR-HR paired images online and trained TATT using these images.
The evaluation involved three datasets: SynTZ, SynSTR, and STR, introduced in Sec. \ref{sec:effectiveness_of_daatset_extension}.
Note that the original TextZoom was not included in these datasets to ensure a fair comparison.
The comparison results are presented in Tab. \ref{table:comparison_to_bsrgan}.
As can be seen, the utilization of paired images generated by our framework results in significantly higher recognition accuracy compared to those generated by the BSRGAN and Real-ESRGAN pipelines.
On the other hand, their SSIM and PSNR scores remain competitive.

\begin{table}[t]
  \centering
  \begin{minipage}[t]{0.45\textwidth}
  \centering
  \resizebox{1\textwidth}{!}{
  \begin{tabular}{c|ccc|ccc}
    \hline
     & \multicolumn{3}{c|}{SSIM ($\times 10^{-2}$)} & \multicolumn{3}{c}{PSNR} \\
    \hline
    Method & SynTZ & SynSTR & STR & SynTZ & SynSTR & STR\\
    \hline
  BSRGAN & 73.05  & 75.50 & 74.56  & \textbf{21.42} & \textbf{21.87} & \textbf{21.36} \\
  Real-ESRGAN & 72.29 & 74.29 & 72.47  & 21.15 & 21.20 & 20.12  \\
   Ours & \textbf{73.95} & \textbf{76.39} & \textbf{75.97}  & 20.40 & 21.15 &  20.66  \\
  \hline
  \end{tabular}
  }
  \subcaption{}
  \end{minipage}
  \\
  \begin{minipage}[t]{0.44\textwidth}
  \centering
  \resizebox{0.65\textwidth}{!}{
  \begin{tabular}{c|ccc}
    \hline
     & \multicolumn{3}{c}{Acc. (\%)} \\
    \hline
    Method & SynTZ & SynSTR & STR \\
    \hline
  BSRGAN & 40.82 & 46.50 & 47.62 \\
  Real-ESRGAN  & 40.36 & 44.50  & 45.03 \\
  Ours & \textbf{48.46} & \textbf{52.69} & \textbf{54.03} \\
  \hline
  \end{tabular}
  }
  \subcaption{}
  \end{minipage}
  
\caption{Comparison with synthetic degradation methods. TATT was trained with these paired images, and the performance was evaluated in terms of average of (a) SSIM/PSNR and (b) recognition accuracy on TextZoom.}
\label{table:comparison_to_bsrgan}
\end{table}

\subsection{KL Loss vs. CTC Loss}
\label{sec:tp_gen_loss}
\begin{figure}[t!]
    \begin{tabular}{c}
      \begin{minipage}[t]{0.95\linewidth}
        \centering
        \includegraphics[width=7.0cm]{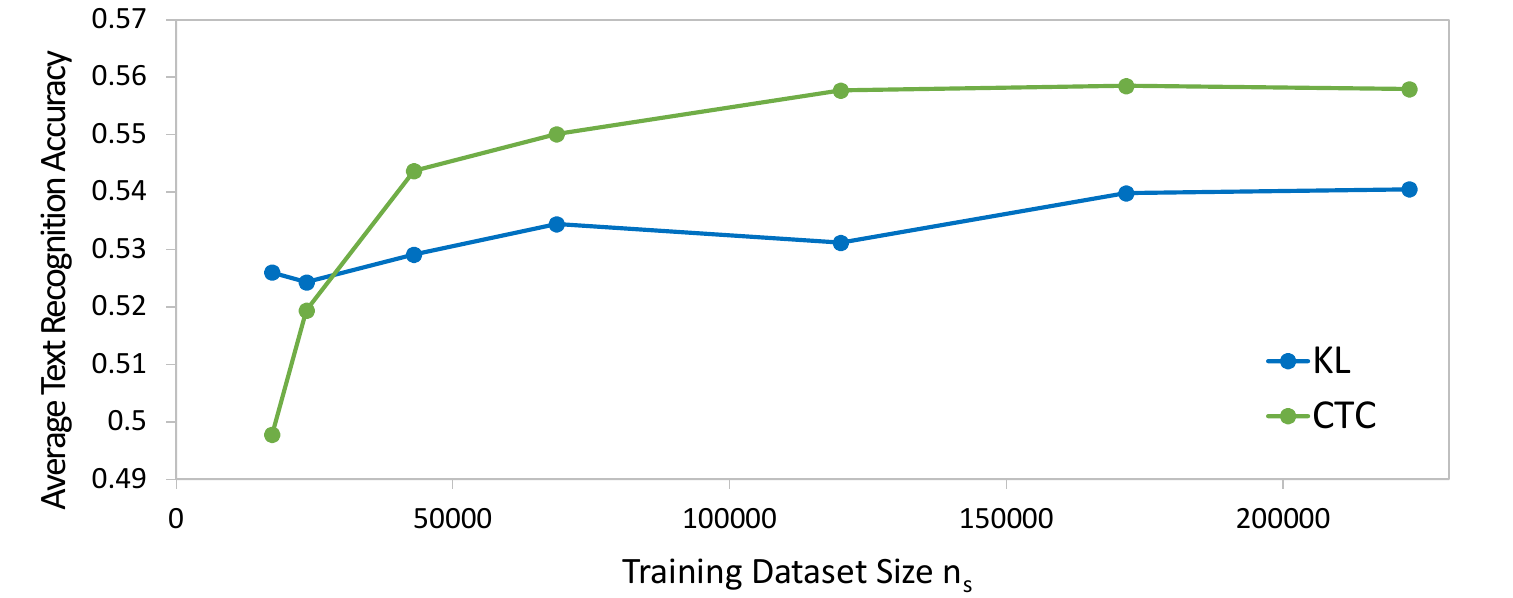}
      \end{minipage}
    \end{tabular}
    \caption{Comparison between KL and CTC losses. It shows the average recognition accuracy of TATT trained with KL and CTC losses for varying training dataset sizes.}
\label{fig:tp_vs_ctc}
\end{figure}

KL loss \cite{ma2021text} and CTC loss \cite{10.1145/1143844.1143891} were used to train the text prior generator using TextZoom and the augmented dataset, respectively.
This is because the appropriate dataset size varies depending on the loss function.
Figure \ref{fig:tp_vs_ctc} compares the recognition accuracy with the KL and CTC losses for different training dataset sizes.
We can see that the original TextZoom is not sufficiently large when CTC loss is used while CTC loss is more effective than KL loss for a larger size of the augmented dataset.
The CTC loss tends to induce overfitting when the dataset size is small because it directly measures the distance from the ground-truth texts.
On the other hand, KL loss measures the distance from the probability predicted for HR images and is less likely to induce overfitting while the distance may not be accurate.

\subsection{Effects of Super-resolver and Degrader}
Synthesizer was trained on HR text images of TextZoom and the preprocessed scene text recognition dataset.
These datasets contain blurred images as well as clear images, resulting in Synthesizer generating both blurred and clear images.
Therefore, without Super-resolver and Degrader, it is anticipated that the proposed framework cannot create stable-quality paired images.
Here, we refer to the images generated by Synthesizer as medium-resolution (MR) images.
To examine the effectiveness of the LR-HR paired images, we created two extended datasets: one without Super-resolver (LR-MR paired images) and the other without Degrader (MR-HR paired images).
We trained TATT on these datasets and compared them with the results of LR-HR paired images.
As shown in Tab. \ref{table:effect_ddsr}, the highest recognition accuracy was achieved when LR-HR paired images were used.
Although the highest SSIM/PSNR were obtained with the LR-MR pairs, we consider that this is due to the domain gap between the synthesized and the original HR images (see Sec. \ref{sec:effectiveness_of_daatset_extension} and \ref{sec:higher-resolution_text_images} for more details).

\begin{table}[t]
  \centering
  \resizebox{0.4\textwidth}{!}{
  \begin{tabular}{c|ccc}
    \hline
    Input-Target & SSIM ($\times 10^{-2}$) & PSNR & Acc. (\%) \\
    \hline
  LR-MR & \textbf{79.08} & \textbf{22.26} & 54.40 \\
  MR-HR & 78.32 & 22.00 & 52.74  \\
  LR-HR & 78.49 & 21.92 & \textbf{55.77} \\
  \hline
  \end{tabular}
    }
\caption{Comparison of LR-HR pairs with two different input-target pairs: LR-MR and MR-HR pairs. TATT was trained on each of the three pairs and the performance was evaluated on the test set of TextZoom.  MR images represent images generated by Synthesizer trained on TextZoom only.}
\label{table:effect_ddsr}
\end{table}

\section{Limitation}
Super-resolver and Degrader must be trained on TextZoom.
Thus, non-simple text images, which are not included in TextZoom such as those shown in Fig. \ref{fig:difficult_samples}, cannot be handled in the proposed framework.
To alleviate this drawback, extensions such as combining with image degradation methods that do not require paired images can be considered.

\begin{figure}[t!]
    \begin{tabular}{c}
      \begin{minipage}[t]{0.95\linewidth}
        \centering
        \includegraphics[width=7.cm]{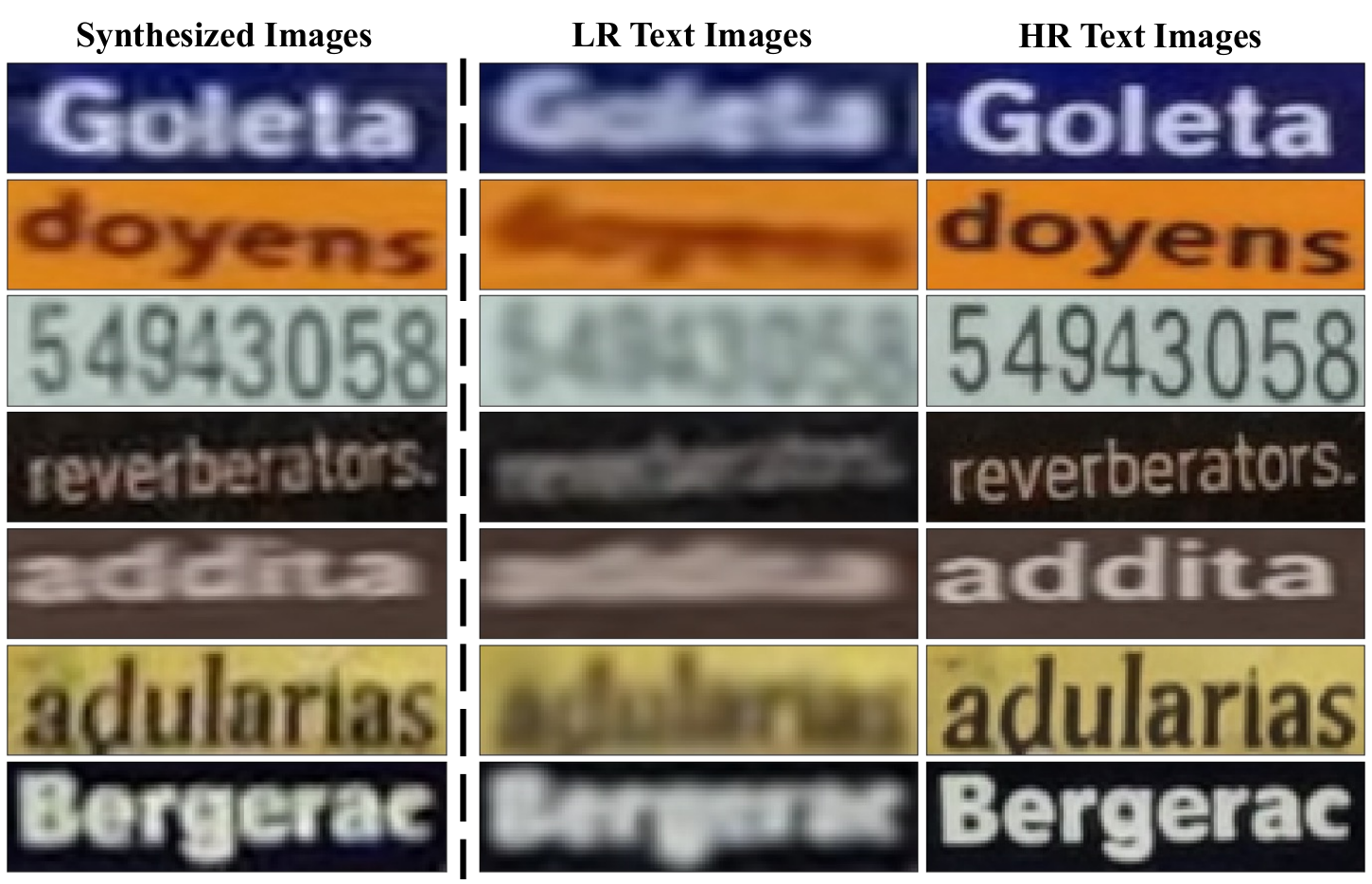}
      \end{minipage}
      
    \end{tabular}
    \caption{Examples of text images generated by the proposed framework. The last two columns show the LR-HR paired images and the first column shows the images generated by Synthesizer.}
\label{fig:lr_hr_examples}
\end{figure}

\section{Conclusions}
We experimentally showed that text-conditional DMs are effective in STISR, achieving state-of-the-art results in the TextZoom evaluation. 
Notably, the performance of the text-conditional DMs becomes remarkably enhanced when trained using ground-truth texts.
Leveraging this exceptionally expressive capacity, we proposed a novel framework for synthesizing LR-HR paired text images.
Our proposed framework, encompassing three distinct text-conditional DMs, can generate high-quality paired text images from user-provided texts.
Our experiments demonstrated a marked improvement in the performance of STISR methods when trained using the paired images generated through our proposed framework.

{\small
\bibliographystyle{ieee_fullname}
\bibliography{egbib}

\begin{thebibliography}{10}\itemsep=-1pt

\bibitem{baek2021if}
Jeonghun Baek, Yusuke Matsui, and Kiyoharu Aizawa.
\newblock What if we only use real datasets for scene text recognition? toward scene text recognition with fewer labels.
\newblock In {\em Proceedings of the IEEE/CVF Conference on Computer Vision and Pattern Recognition}, pages 3113--3122, 2021.

\bibitem{baranchuk2022labelefficient}
Dmitry Baranchuk, Andrey Voynov, Ivan Rubachev, Valentin Khrulkov, and Artem Babenko.
\newblock Label-efficient semantic segmentation with diffusion models.
\newblock In {\em Proceedings of the International Conference on Learning Representations}, 2022.

\bibitem{10.1007/978-3-031-19815-1_11}
Darwin Bautista and Rowel Atienza.
\newblock Scene text recognition with permuted autoregressive sequence models.
\newblock In {\em Proceedings of the European Conference on Computer Vision}, page 178–196, 2022.

\bibitem{10.3389/frsip.2023.1106465}
Ahmed Cheikh~Sidiya, Xuan Xu, Ning Xu, and Xin Li.
\newblock Degradation learning and skip-transformer for blind face restoration.
\newblock {\em Frontiers in Signal Processing}, 3, 2023.

\bibitem{Chen_2021_CVPR}
Jingye Chen, Bin Li, and Xiangyang Xue.
\newblock Scene text telescope: Text-focused scene image super-resolution.
\newblock In {\em Proceedings of the IEEE/CVF Conference on Computer Vision and Pattern Recognition}, pages 12026--12035, June 2021.

\bibitem{chen2022text}
Jingye Chen, Haiyang Yu, Jianqi Ma, Bin Li, and Xiangyang Xue.
\newblock Text gestalt: Stroke-aware scene text image super-resolution.
\newblock In {\em Proceedings of the AAAI Conference on Artificial Intelligence}, volume~36, pages 285--293, 2022.

\bibitem{chng2019icdar2019}
Chee~Kheng Chng, Yuliang Liu, Yipeng Sun, Chun~Chet Ng, Canjie Luo, Zihan Ni, ChuanMing Fang, Shuaitao Zhang, Junyu Han, Errui Ding, et~al.
\newblock Icdar2019 robust reading challenge on arbitrary-shaped text-rrc-art.
\newblock In {\em Proceedings of the IEEE International Conference on Document Analysis and Recognition}, pages 1571--1576, 2019.

\bibitem{croitoru2022diffusion}
Florinel-Alin Croitoru, Vlad Hondru, Radu~Tudor Ionescu, and Mubarak Shah.
\newblock Diffusion models in vision: A survey.
\newblock {\em arXiv preprint arXiv:2209.04747}, 2022.

\bibitem{NEURIPS2021_49ad23d1}
Prafulla Dhariwal and Alexander Nichol.
\newblock Diffusion models beat gans on image synthesis.
\newblock In {\em Advances in Neural Information Processing Systems}, volume~34, 2021.

\bibitem{dong2015image}
Chao Dong, Chen~Change Loy, Kaiming He, and Xiaoou Tang.
\newblock Image super-resolution using deep convolutional networks.
\newblock {\em IEEE transactions on pattern analysis and machine intelligence}, 38(2):295--307, 2015.

\bibitem{dong2015boosting}
Chao Dong, Ximei Zhu, Yubin Deng, Chen~Change Loy, and Yu Qiao.
\newblock Boosting optical character recognition: A super-resolution approach.
\newblock {\em arXiv preprint arXiv:1506.02211}, 2015.

\bibitem{Fang2021ReadLH}
Shancheng Fang, Hongtao Xie, Yuxin Wang, Zhendong Mao, and Yongdong Zhang.
\newblock Read like humans: Autonomous, bidirectional and iterative language modeling for scene text recognition.
\newblock In {\em Proceedings of the IEEE/CVF Conference on Computer Vision and Pattern Recognition}, pages 7094--7103, 2021.

\bibitem{10.1145/1143844.1143891}
Alex Graves, Santiago Fern\'{a}ndez, Faustino Gomez, and J\"{u}rgen Schmidhuber.
\newblock Connectionist temporal classification: Labelling unsegmented sequence data with recurrent neural networks.
\newblock In {\em Proceedings of the International Conference on Machine Learning}, page 369–376, 2006.

\bibitem{gupta2016synthetic}
Ankush Gupta, Andrea Vedaldi, and Andrew Zisserman.
\newblock Synthetic data for text localisation in natural images.
\newblock In {\em Proceedings of the IEEE conference on computer vision and pattern recognition}, pages 2315--2324, 2016.

\bibitem{ho2020denoising}
Jonathan Ho, Ajay Jain, and Pieter Abbeel.
\newblock Denoising diffusion probabilistic models.
\newblock {\em Advances in Neural Information Processing Systems}, 33:6840--6851, 2020.

\bibitem{ho2022cascaded}
Jonathan Ho, Chitwan Saharia, William Chan, David~J Fleet, Mohammad Norouzi, and Tim Salimans.
\newblock Cascaded diffusion models for high fidelity image generation.
\newblock {\em Journal of Machine Learning Research}, 23:47--1, 2022.

\bibitem{ho2021classifierfree}
Jonathan Ho and Tim Salimans.
\newblock Classifier-free diffusion guidance.
\newblock In {\em NeurIPS Workshop on Deep Generative Models and Downstream Applications}, 2021.

\bibitem{mjsynth}
Max Jaderberg, Karen Simonyan, Andrea Vedaldi, and Andrew Zisserman.
\newblock Synthetic data and artificial neural networks for natural scene text recognition.
\newblock In {\em NeurIPS Workshop on Deep Learning}, 2014.

\bibitem{10.1145/3528223.3530104}
Yuming Jiang, Shuai Yang, Haonan Qju, Wayne Wu, Chen~Change Loy, and Ziwei Liu.
\newblock Text2human: Text-driven controllable human image generation.
\newblock {\em ACM Transactions on Graphics}, 41(4), 2022.

\bibitem{7333942}
Dimosthenis Karatzas, Lluis Gomez-Bigorda, Anguelos Nicolaou, Suman Ghosh, Andrew Bagdanov, Masakazu Iwamura, Jiri Matas, Lukas Neumann, Vijay~Ramaseshan Chandrasekhar, Shijian Lu, Faisal Shafait, Seiichi Uchida, and Ernest Valveny.
\newblock Icdar competition on robust reading.
\newblock In {\em Proceedings of 13th International Conference on Document Analysis and Recognition}, pages 1156--1160, 2015.

\bibitem{6628859}
Dimosthenis Karatzas, Faisal Shafait, Seiichi Uchida, Masakazu Iwamura, Lluis Gomez~i Bigorda, Sergi~Robles Mestre, Joan Mas, David~Fernandez Mota, Jon~Almazàn Almazàn, and Lluís~Pere de~las Heras.
\newblock Icdar competition on robust reading.
\newblock In {\em Proceedings of 12th International Conference on Document Analysis and Recognition}, pages 1484--1493, 2013.

\bibitem{karras2020analyzing}
Tero Karras, Samuli Laine, Miika Aittala, Janne Hellsten, Jaakko Lehtinen, and Timo Aila.
\newblock Analyzing and improving the image quality of stylegan.
\newblock In {\em Proceedings of the IEEE/CVF Conference on Computer Vision and Pattern Recognition}, pages 8110--8119, 2020.

\bibitem{7780551}
Jiwon Kim, Jung~Kwon Lee, and Kyoung~Mu Lee.
\newblock Accurate image super-resolution using very deep convolutional networks.
\newblock In {\em Proceedings of the IEEE Conference on Computer Vision and Pattern Recognition}, pages 1646--1654, 2016.

\bibitem{lai2017deep}
Wei-Sheng Lai, Jia-Bin Huang, Narendra Ahuja, and Ming-Hsuan Yang.
\newblock Deep laplacian pyramid networks for fast and accurate super-resolution.
\newblock In {\em Proceedings of the IEEE conference on computer vision and pattern recognition}, pages 624--632, 2017.

\bibitem{Ledig_2017_CVPR}
Christian Ledig, Lucas Theis, Ferenc Huszar, Jose Caballero, Andrew Cunningham, Alejandro Acosta, Andrew Aitken, Alykhan Tejani, Johannes Totz, Zehan Wang, and Wenzhe Shi.
\newblock Photo-realistic single image super-resolution using a generative adversarial network.
\newblock In {\em Proceedings of the IEEE Conference on Computer Vision and Pattern Recognition}, July 2017.

\bibitem{lim2017enhanced}
Bee Lim, Sanghyun Son, Heewon Kim, Seungjun Nah, and Kyoung Mu~Lee.
\newblock Enhanced deep residual networks for single image super-resolution.
\newblock In {\em Proceedings of the IEEE conference on computer vision and pattern recognition workshops}, pages 136--144, 2017.

\bibitem{LUO2019109}
Canjie Luo, Lianwen Jin, and Zenghui Sun.
\newblock Moran: A multi-object rectified attention network for scene text recognition.
\newblock {\em Pattern Recognition}, 90:109--118, 2019.

\bibitem{ma2021text}
Jianqi Ma, Shi Guo, and Lei Zhang.
\newblock Text prior guided scene text image super-resolution.
\newblock {\em arXiv preprint arXiv:2106.15368}, 2021.

\bibitem{Ma_2022_CVPR}
Jianqi Ma, Zhetong Liang, and Lei Zhang.
\newblock A text attention network for spatial deformation robust scene text image super-resolution.
\newblock In {\em Proceedings of the IEEE/CVF Conference on Computer Vision and Pattern Recognition}, pages 5911--5920, June 2022.

\bibitem{maeda2020unpaired}
Shunta Maeda.
\newblock Unpaired image super-resolution using pseudo-supervision.
\newblock In {\em Proceedings of the IEEE/CVF Conference on Computer Vision and Pattern Recognition}, pages 291--300, 2020.

\bibitem{Mishra2009SceneTR}
Anand Mishra, Alahari Karteek, and C.~V. Jawahar.
\newblock Scene text recognition using higher order language priors.
\newblock In {\em Proceedings of the British Machine Vision Conference}, 2009.

\bibitem{10.1007/978-3-030-58555-6_10}
Yongqiang Mou, Lei Tan, Hui Yang, Jingying Chen, Leyuan Liu, Rui Yan, and Yaohong Huang.
\newblock Plugnet: Degradation aware scene text recognition supervised by a pluggable super-resolution unit.
\newblock In {\em Proceedings of the European Conference on Computer Vision}, pages 158--174, 2020.

\bibitem{nayef2019icdar2019}
Nibal Nayef, Yash Patel, Michal Busta, Pinaki~Nath Chowdhury, Dimosthenis Karatzas, Wafa Khlif, Jiri Matas, Umapada Pal, Jean-Christophe Burie, Cheng-lin Liu, et~al.
\newblock Icdar2019 robust reading challenge on multi-lingual scene text detection and recognition—rrc-mlt-2019.
\newblock In {\em Proceedings of the IEEE International conference on document analysis and recognition}, pages 1582--1587, 2019.

\bibitem{nichol2021glide}
Alex Nichol, Prafulla Dhariwal, Aditya Ramesh, Pranav Shyam, Pamela Mishkin, Bob McGrew, Ilya Sutskever, and Mark Chen.
\newblock Glide: Towards photorealistic image generation and editing with text-guided diffusion models.
\newblock In {\em Proceedings of the International Conference on Machine Learning}, volume 162, 2022.

\bibitem{pmlr-v139-nichol21a}
Alexander~Quinn Nichol and Prafulla Dhariwal.
\newblock Improved denoising diffusion probabilistic models.
\newblock In {\em Proceedings of the International Conference on Machine Learning}, volume 139, pages 8162--8171, 2021.

\bibitem{9040515}
Yuhui Quan, Jieting Yang, Yixin Chen, Yong Xu, and Hui Ji.
\newblock Collaborative deep learning for super-resolving blurry text images.
\newblock {\em IEEE Transactions on Computational Imaging}, 6:778--790, 2020.

\bibitem{ramesh2022hierarchical}
Aditya Ramesh, Prafulla Dhariwal, Alex Nichol, Casey Chu, and Mark Chen.
\newblock Hierarchical text-conditional image generation with clip latents.
\newblock {\em arXiv preprint arXiv:2204.06125}, 2022.

\bibitem{Rombach_2022_CVPR}
Robin Rombach, Andreas Blattmann, Dominik Lorenz, Patrick Esser, and Bj\"orn Ommer.
\newblock High-resolution image synthesis with latent diffusion models.
\newblock In {\em Proceedings of the IEEE/CVF Conference on Computer Vision and Pattern Recognition}, pages 10684--10695, June 2022.

\bibitem{10.1007/978-3-319-24574-4_28}
Olaf Ronneberger, Philipp Fischer, and Thomas Brox.
\newblock U-net: Convolutional networks for biomedical image segmentation.
\newblock In {\em Proceedings of the Medical Image Computing and Computer-Assisted Intervention}, pages 234--241, 2015.

\bibitem{saharia2022photorealistic}
Chitwan Saharia, William Chan, Saurabh Saxena, Lala Li, Jay Whang, Emily Denton, Seyed Kamyar~Seyed Ghasemipour, Burcu~Karagol Ayan, S~Sara Mahdavi, Rapha~Gontijo Lopes, et~al.
\newblock Photorealistic text-to-image diffusion models with deep language understanding.
\newblock {\em arXiv preprint arXiv:2205.11487}, 2022.

\bibitem{9887996}
Chitwan Saharia, Jonathan Ho, William Chan, Tim Salimans, David~J. Fleet, and Mohammad Norouzi.
\newblock Image super-resolution via iterative refinement.
\newblock {\em IEEE Transactions on Pattern Analysis and Machine Intelligence}, pages 1--14, 2022.

\bibitem{7801919}
Baoguang Shi, Xiang Bai, and Cong Yao.
\newblock An end-to-end trainable neural network for image-based sequence recognition and its application to scene text recognition.
\newblock {\em IEEE Transactions on Pattern Analysis and Machine Intelligence}, 39(11):2298--2304, 2017.

\bibitem{8395027}
Baoguang Shi, Mingkun Yang, Xinggang Wang, Pengyuan Lyu, Cong Yao, and Xiang Bai.
\newblock Aster: An attentional scene text recognizer with flexible rectification.
\newblock {\em IEEE Transactions on Pattern Analysis and Machine Intelligence}, 41(9):2035--2048, 2019.

\bibitem{shi2017icdar2017}
Baoguang Shi, Cong Yao, Minghui Liao, Mingkun Yang, Pei Xu, Linyan Cui, Serge Belongie, Shijian Lu, and Xiang Bai.
\newblock Icdar2017 competition on reading chinese text in the wild (rctw-17).
\newblock In {\em Proceedings of the IEEE international conference on document analysis and recognition}, volume~1, pages 1429--1434, 2017.

\bibitem{pmlr-v37-sohl-dickstein15}
Jascha Sohl-Dickstein, Eric Weiss, Niru Maheswaranathan, and Surya Ganguli.
\newblock Deep unsupervised learning using nonequilibrium thermodynamics.
\newblock In {\em Proceedings of the International Conference on Machine Learning}, volume~37, pages 2256--2265, 2015.

\bibitem{song2019generative}
Yang Song and Stefano Ermon.
\newblock Generative modeling by estimating gradients of the data distribution.
\newblock {\em Advances in Neural Information Processing Systems}, 32, 2019.

\bibitem{sun2019icdar}
Yipeng Sun, Zihan Ni, Chee-Kheng Chng, Yuliang Liu, Canjie Luo, Chun~Chet Ng, Junyu Han, Errui Ding, Jingtuo Liu, Dimosthenis Karatzas, et~al.
\newblock Icdar 2019 competition on large-scale street view text with partial labeling-rrc-lsvt.
\newblock In {\em Proceedings of the IEEE International Conference on Document Analysis and Recognition}, pages 1557--1562, 2019.

\bibitem{NIPS2017_3f5ee243}
Ashish Vaswani, Noam Shazeer, Niki Parmar, Jakob Uszkoreit, Llion Jones, Aidan~N Gomez, \L~ukasz Kaiser, and Illia Polosukhin.
\newblock Attention is all you need.
\newblock In {\em Advances in Neural Information Processing Systems}, volume~30, 2017.

\bibitem{veit2016coco}
Andreas Veit, Tomas Matera, Lukas Neumann, Jiri Matas, and Serge Belongie.
\newblock Coco-text: Dataset and benchmark for text detection and recognition in natural images.
\newblock {\em arXiv preprint arXiv:1601.07140}, 2016.

\bibitem{6126402}
Kai Wang, Boris Babenko, and Serge Belongie.
\newblock End-to-end scene text recognition.
\newblock In {\em Proceedings of the IEEE International Conference on Computer Vision}, pages 1457--1464, 2011.

\bibitem{10.1007/978-3-030-58607-2_38}
Wenjia Wang, Enze Xie, Xuebo Liu, Wenhai Wang, Ding Liang, Chunhua Shen, and Xiang Bai.
\newblock Scene text image super-resolution in the wild.
\newblock In {\em Proceedings of the European Conference on Computer Vision}, pages 650--666, 2020.

\bibitem{DBLP:journals/corr/abs-1909-07113}
Wenjia Wang, Enze Xie, Peize Sun, Wenhai Wang, Lixun Tian, Chunhua Shen, and Ping Luo.
\newblock Textsr: Content-aware text super-resolution guided by recognition.
\newblock {\em CoRR}, abs/1909.07113, 2019.

\bibitem{wang2021real}
Xintao Wang, Liangbin Xie, Chao Dong, and Ying Shan.
\newblock Real-esrgan: Training real-world blind super-resolution with pure synthetic data.
\newblock In {\em Proceedings of the IEEE/CVF International Conference on Computer Vision Workshops}, pages 1905--1914, 2021.

\bibitem{8237298}
Xiangyu Xu, Deqing Sun, Jinshan Pan, Yujin Zhang, Hanspeter Pfister, and Ming-Hsuan Yang.
\newblock Learning to super-resolve blurry face and text images.
\newblock In {\em Proceedings of the IEEE International Conference on Computer Vision}, pages 251--260, 2017.

\bibitem{yang2022diffusion}
Ling Yang, Zhilong Zhang, Yang Song, Shenda Hong, Runsheng Xu, Yue Zhao, Yingxia Shao, Wentao Zhang, Bin Cui, and Ming-Hsuan Yang.
\newblock Diffusion models: A comprehensive survey of methods and applications.
\newblock {\em arXiv preprint arXiv:2209.00796}, 2022.

\bibitem{zhang2021designing}
Kai Zhang, Jingyun Liang, Luc Van~Gool, and Radu Timofte.
\newblock Designing a practical degradation model for deep blind image super-resolution.
\newblock In {\em Proceedings of the IEEE/CVF International Conference on Computer Vision}, pages 4791--4800, 2021.

\bibitem{zhang2019icdar}
Rui Zhang, Yongsheng Zhou, Qianyi Jiang, Qi Song, Nan Li, Kai Zhou, Lei Wang, Dong Wang, Minghui Liao, Mingkun Yang, et~al.
\newblock Icdar 2019 robust reading challenge on reading chinese text on signboard.
\newblock In {\em Proceedings of the IEEE International conference on document analysis and recognition}, pages 1577--1581, 2019.

\bibitem{zhang2017uber}
Ying Zhang, Lionel Gueguen, Ilya Zharkov, Peter Zhang, Keith Seifert, and Ben Kadlec.
\newblock Uber-text: A large-scale dataset for optical character recognition from street-level imagery.
\newblock In {\em Proceedings of the IEEE conference on computer vision and pattern recognition workshops}, volume 2017, page~5, 2017.

\bibitem{zhang2019multiple}
Yongbing Zhang, Siyuan Liu, Chao Dong, Xinfeng Zhang, and Yuan Yuan.
\newblock Multiple cycle-in-cycle generative adversarial networks for unsupervised image super-resolution.
\newblock {\em IEEE transactions on Image Processing}, 29:1101--1112, 2019.

\bibitem{Zhang_2018_CVPR}
Yulun Zhang, Yapeng Tian, Yu Kong, Bineng Zhong, and Yun Fu.
\newblock Residual dense network for image super-resolution.
\newblock In {\em Proceedings of the IEEE Conference on Computer Vision and Pattern Recognition}, June 2018.

\bibitem{10.1145/3474085.3475469}
Cairong Zhao, Shuyang Feng, Brian~Nlong Zhao, Zhijun Ding, Jun Wu, Fumin Shen, and Heng~Tao Shen.
\newblock Scene text image super-resolution via parallelly contextual attention network.
\newblock In {\em Proceedings of the ACM International Conference on Multimedia}, page 2908–2917, 2021.

\bibitem{zhao2022c3}
Minyi Zhao, Miao Wang, Fan Bai, Bingjia Li, Jie Wang, and Shuigeng Zhou.
\newblock C3-stisr: Scene text image super-resolution with triple clues.
\newblock In {\em Proceedings of the International Joint Conferences on Artificial Intelligence}, pages 1707--1713, 2022.

\bibitem{zhu2017unpaired}
Jun-Yan Zhu, Taesung Park, Phillip Isola, and Alexei~A Efros.
\newblock Unpaired image-to-image translation using cycle-consistent adversarial networks.
\newblock In {\em Proceedings of the IEEE International Conference on Computer Vision}, pages 2223--2232, 2017.

\end{thebibliography}
}

\clearpage

\appendix

\section{Denoising Diffusion Probabilistic Models}
\label{sec:ddpm}
A diffusion model consists of two processes, forward and reverse.
In the forward process, Gaussian noise is gradually added to the input image and, eventually, becomes pure Gaussian noise.
Conversely, in the reverse process, starting from pure Gaussian noise, noise is removed sequentially to recreate the original images.
Following this definition, DMs can be classified into at least three categories \cite{croitoru2022diffusion,yang2022diffusion}: denoising diffusion probabilistic models and noise-conditioned score networks, and stochastic differential equations.
The method proposed in this study is based on DDPMs.
In the following, we derive the loss function of DDPMs in Eq. \ref{eq:conditional_ddpm_loss}.
The following derivation is based on \cite{pmlr-v37-sohl-dickstein15,ho2020denoising}.

Given a data distribution ${x}_0\sim q({x_0})$, the forward process is defined as the Markov process, where a series of latent variables $x_1,\dots,x_T$ is produced by progressively adding Gaussian noise:

\begin{equation}
    q(x_t|x_{t-1}) = \mathcal{N}(\sqrt{1-\beta_t}x_{t-1},\beta_t\boldsymbol{I}),
    \label{eq:q_xt_xt_1}
\end{equation}
to the sample. Hence,
\begin{equation}
    q(x_{1:T}|x_0)=\prod_{t=1}^Tq(x_t|x_{t-1}).
    \label{eq:qx1t_x0}
\end{equation}
Here, $\beta_t\in(0,1)$, $\forall t\in\{1,\dots,T\}$ indicates the variance at time $t$.
When $T$ is sufficiently large, $x_T$ is equivalent to a pure Gaussian noise. 
Here, there is a helpful property in that $x_t$ at any timestep $t$ can be sampled directly from $x_0$, following a single Gaussian

\begin{equation}
    q(x_t|x_0) = \mathcal{N}(\sqrt{\Bar{\alpha}_t}x_0, (1-\Bar{\alpha}_t)\boldsymbol{I}),
    \label{eq:q_xt_x0}
\end{equation}
where $\alpha_t=1-\beta_t$ and $\Bar{\alpha}_t=\prod_{s=0}^t\alpha_t$.
Therefore, using $\epsilon\sim\mathcal{N}(0,\boldsymbol{I})$, $x_t=\sqrt{\Bar{\alpha}_t}x_0+(1-\Bar{\alpha}_t)\epsilon$.

In the reverse process, starting from a Gaussian noise $x_T\sim\mathcal{N}(0,\boldsymbol{I})$, we can reverse the forward process by sampling from the posterior $q(x_{t-1}|x_t)$.
However, directly estimating $q(x_{t-1}|x_t)$ is difficult because it requires the data distribution $q(x_0)$.
However, it is known that $q(x_{t-1}|x_t)$ can be approximated as a Gaussian distribution when $\beta_t$ is sufficiently small \cite{pmlr-v37-sohl-dickstein15}.
Therefore, $q(x_{t-1}|x_t)$ can reasonably fit to the true posterior by being parameterized as

\begin{equation}
    p_\theta(x_{t-1}|x_t) = \mathcal{N}(\mu_\theta(x_t,t),\Sigma_\theta(x_t,t)).
    \label{eq:p_xt_1_xt}
\end{equation}
Consequently, the reverse process is parameterized as follows:
\begin{equation}
    p_{\theta}(x_{0:T})=p(x_T)\prod_{t=1}^{T} p_\theta(x_{t-1}|x_t)
    \label{eq:ptheta_xT}
\end{equation}

The loss function to be optimized is provided by the variational lower bound of the negative log likelihood $\mathcal{L}=\mathbb{E}[-\log p_\theta(x_0)]$.
This optimization can be performed efficiently for each timestep because $\mathcal{L}$ can be broken down by using chain rules and Eq. \ref{eq:q_xt_x0}.
The final loss function is obtained as follows:

\begin{equation}
    \mathcal{L}_{u}=\mathbb{E}_{t,x_0,\epsilon}[\|\epsilon-\epsilon_\theta(x_t,t)\|^2].
    \label{eq:unconditional_ddpm_loss_app}
\end{equation}
See \cite{pmlr-v37-sohl-dickstein15,ho2020denoising} for the detailed derivation of Eq. \ref{eq:unconditional_ddpm_loss_app}.
Consequently, a deep neural network $\epsilon_\theta$ is trained to estimate the Gaussian noise $\epsilon$ contained in $x_t$.

\begin{figure}[t!]
    \begin{tabular}{c}
      \begin{minipage}[t]{0.95\linewidth}
        \centering
        \includegraphics[width=8.0cm]{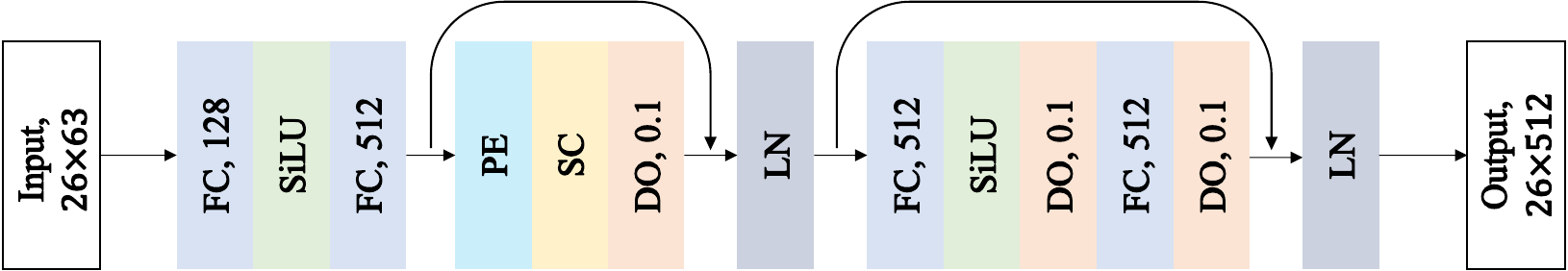}
        \subcaption{}
        \label{fig:text_encoder_architecture_gt_dimss}
      \end{minipage}
       \\
      \begin{minipage}[t]{0.95\linewidth}
        \centering
        \includegraphics[width=8.0cm]{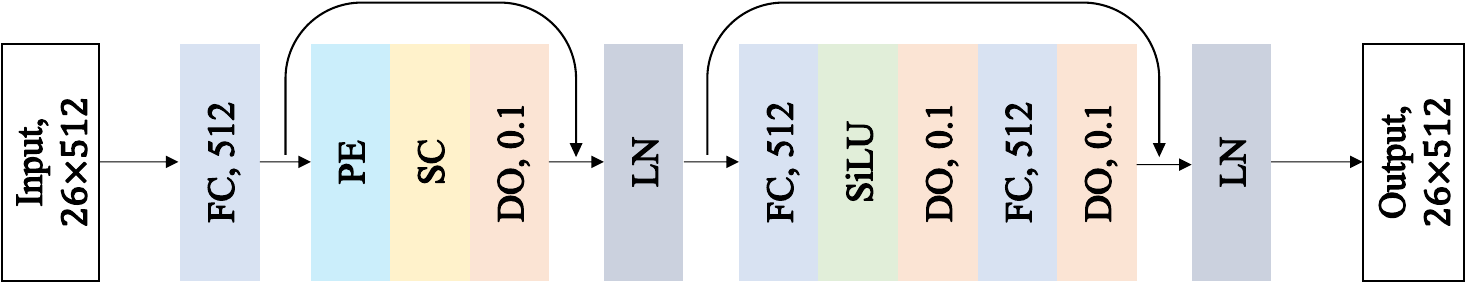}
        \subcaption{}
        \label{fig:text_encoder_architecture_dimss}
      \end{minipage}
    \end{tabular}
    \caption{Architecture details of the text encoders in (a) TCDM and (b) TCDM*, respectively. FC, SA, PE, DO, and LN stand for a fully-connected layer, a self-attention layer, positional encoding, dropout, and layer normalization, respectively.}
\label{fig:text_encoder_architecture}
\end{figure}

\begin{table*}[t!]
  \centering
  \resizebox{0.975\textwidth}{!}{
    \begin{tabular}{c|c|c|c}
        \hline
          Hyperparameters & Traning on TextZoom (Sec. \ref{sec:evaluation_on_stisr}) & Training on Extended dataset (Sec. \ref{sec:evalution_on_extended_dataset}) & Fine-tuning (Sec. \ref{sec:evalution_on_extended_dataset}) \\
        \hline
       Diffusion Steps & 1000 & 1000 & 1000 \\
       Noies Schedule & linear & linear & linear \\
       Channels & 128 & 128 & 128 \\
       Channel Multiplier & 1,2,3,4 & 1,2,3,4 & 1,2,3,4 \\
       Number of Heads & 1 & 1 & 1 \\
       Batch Size & 128 & 128 & 128 \\
       Learning Rate & 3e-4 & 3e-4 & 3e-5  \\
       Iterations & 100K & 300K & 30K \\
       Embedding Dimension & 512 & 512 & 512 \\
       Attention Resolution & 32,16,8 & 32,16,8 & 32,16,8 \\
      \hline
      \end{tabular}
    }
\caption{Hyperparameters of TCDM and TCDM* for each experiment in the main text.}
\label{table:hyperparameters}
\end{table*}

\section{Design of Text-conditional DM Architectures}
\label{sec:design_of_dimss_arc}

To describe the architectural differences simply, we introduce two parameters: $k_b$ and $k_m$.
$k_b$ denotes the number of iterations of the cross-attention and self-attention modules at the bottom of the UNet and $k_m$ denotes the number of iterations in each resolution layer, except for the bottom layer.
Using $k_b$ and $k_m$, the architectures proposed in prior studies \cite{saharia2022photorealistic,Rombach_2022_CVPR} can be expressed as $k_b=1$ and $k_m=1$.
To determine $k_b$ of our text-conditional DM, we used grid search, and Table \ref{table:kb_gt_dimss} shows recognition accuracy of our text-conditional DMs for $k_b\in\{0,3,6,9\}$. TCDM denotes the text-conditional DM and TCDM* indicates TCDM trained using ground-truth text input.
TCDM achieved the best performance when $k_b=6$ and TCDM* when $k_b=9$.
In addition, we set $k_m$ to 0 for TCDM because $k_m>0$ significantly increases the computational cost.
Table \ref{table:text_embedding_methods} shows the inference time per image.
We can see that setting $k_m$ to 1 not only results in a substantial increase in the inference time but also leads to a drop in recognition accuracy.

\begin{table}[t!]
   \centering
  \begin{tabular}{c|cccc}
    \hline
     $k_b$ & 0 & 3 & 6 & 9 \\
    \hline
   TCDM & 55.0\% & 55.1\%  & \textbf{55.7\%}  & 55.6\%  \\
  TCDM* & 55.0\% & 57.5\%  & 67.0\%  & \textbf{68.1\%}  \\
  \hline
  \end{tabular}
  \caption{Average recognition accuracy of TCDM and TCDM* for $k_b\in\{0,3,6,9\}$.}
  \label{table:kb_gt_dimss}
\end{table}

\begin{table}[t]
  \centering
  \begin{tabular}{cc|c|c}
    \hline
    $k_b$ & $k_m$ & Acc. (\%) & Time (s)  \\
    \hline
  1 & 1 & 58.0 & 9.30 \\
  3 & 0  & 57.5 & 5.39 \\
  6 & 0  & 67.0 & 6.07 \\
  9 & 0 & \textbf{68.1} & 6.82 \\
    \hline
  \end{tabular}
\caption{Average recognition accuracy and inference time of TCDM* for $(k_b,k_m)\in\{(1,1),(3,0),(6,0),(9,0)\}$}
\label{table:text_embedding_methods}
\end{table}

The text encoder of our text-conditional DM has a simple architecture, which is based on self-attention and fully-connected layers.
This architecture is based on the text prior generator proposed in \cite{ma2021text}.
Figures \ref{fig:text_encoder_architecture_gt_dimss} and \ref{fig:text_encoder_architecture_dimss} show the architectural details of the text encoders for our text-conditional DMs.
When ground-truth texts are used, the input of the text encoder has a shape $26\times 63$, where the maximum word length $l=26$, and the alphabet size $|\mathcal{A}|=63$.
The input text was encoded into text features for each character candidate, and the output tensor had a shape $26\times 512$.
When the text prior generator is used, the text encoder has the same architecture, except that the input is given as intermediate features of the text prior generator and has a shape $26\times 512$.

The implementation of our text-conditional DMs was based on the code released by the authors in \cite{pmlr-v139-nichol21a,NEURIPS2021_49ad23d1}. In Tab. \ref{table:hyperparameters}, we present hyperparameters of our text-conditional DMs for each experiment in the main text.

\section{Evaluation of Extended Dataset using PSNR}
\label{sec:psnr_evaluation}
\begin{figure}[h!]
    \begin{tabular}{c}
      \begin{minipage}[t]{0.95\linewidth}
        \centering
        \includegraphics[width=8.0cm]{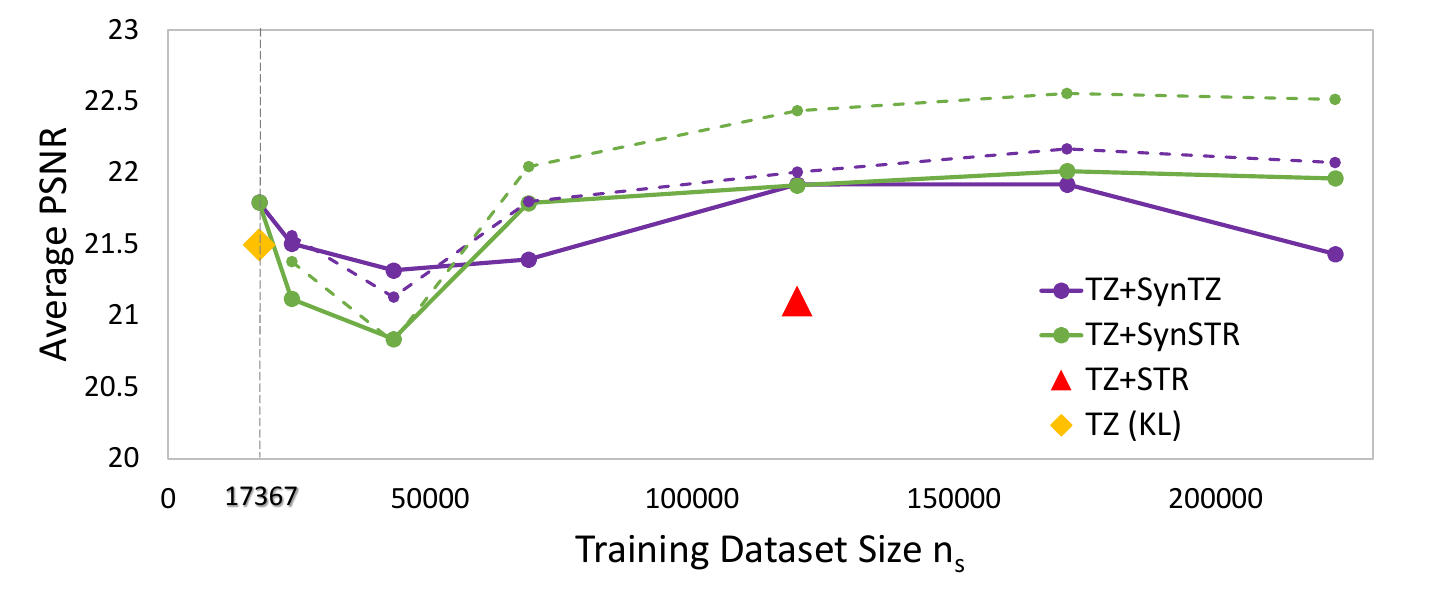}
        \subcaption{}
        \label{fig:psnr_dataset_size_tatt}
      \end{minipage}
        \\
      \begin{minipage}[t]{0.95\linewidth}
        \centering
        \includegraphics[width=8.0cm]{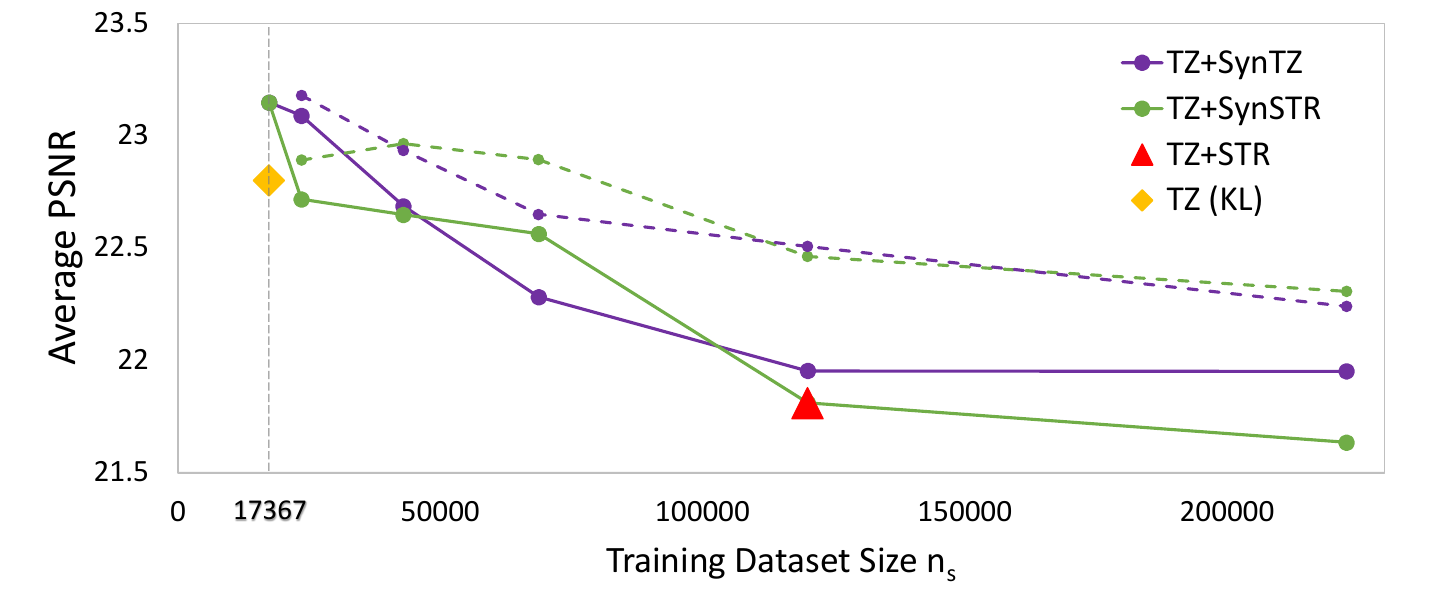}
        \subcaption{}
        \label{fig:psnr_dataset_size_dimss}
      \end{minipage}
    \end{tabular}
    \caption{Evaluation results of (a) TATT and (b) our text-conditional DMs trained on the augmented dataset in terms of PSNR.}
\label{fig:psnr_dataset_size}
\end{figure}

Figures \ref{fig:psnr_dataset_size_tatt} and \ref{fig:psnr_dataset_size_dimss} show the average PSNR of TATT and the text-conditional DMs, respectively, as solid lines.
Similar to the results of SSIM, PSNR does not depend on $n_s$.
Figure \ref{fig:psnr_dataset_size} also shows the results of the fine-tuning with dotted lines.
We can see that PSNR can be significantly improved by fine-tuning.

% TATT (acc)
\begin{figure*}[h!]
    \begin{tabular}{c}
      \begin{minipage}[t]{0.475\linewidth}
        \centering
        \includegraphics[width=8.0cm]{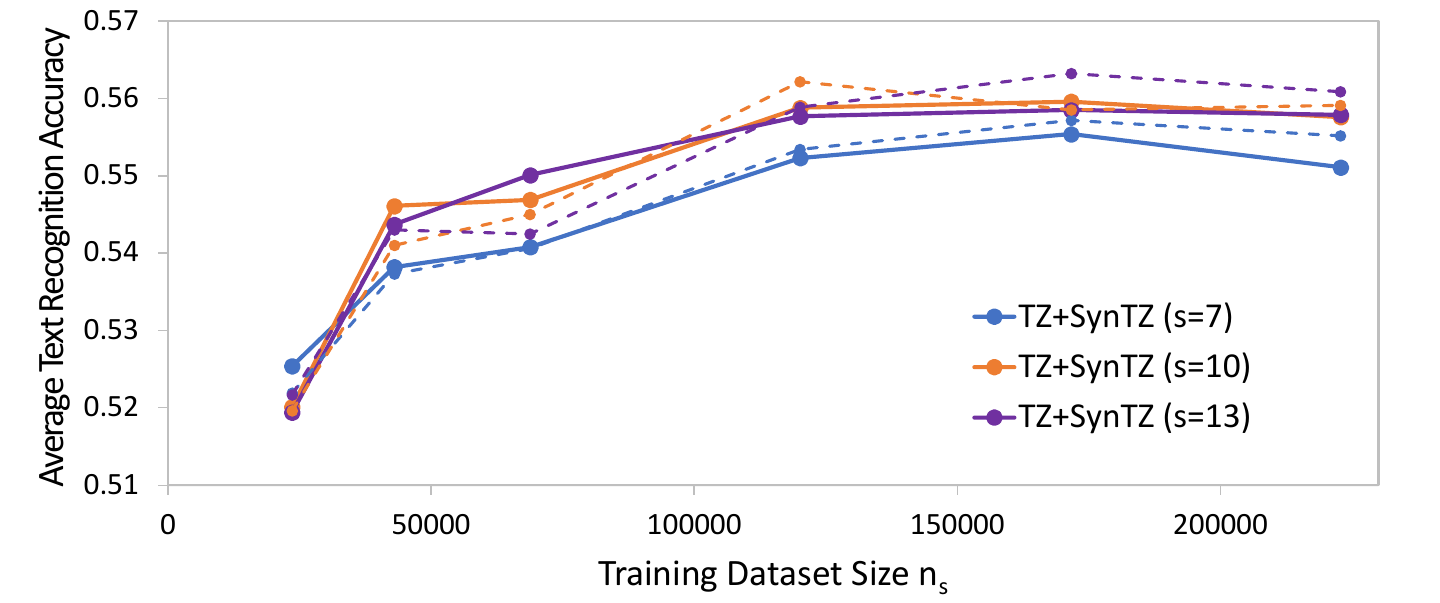}
        \subcaption{}
        \label{fig:acc_max_len_synt}
      \end{minipage}
    
      \begin{minipage}[t]{0.475\linewidth}
        \centering
        \includegraphics[width=8.0cm]{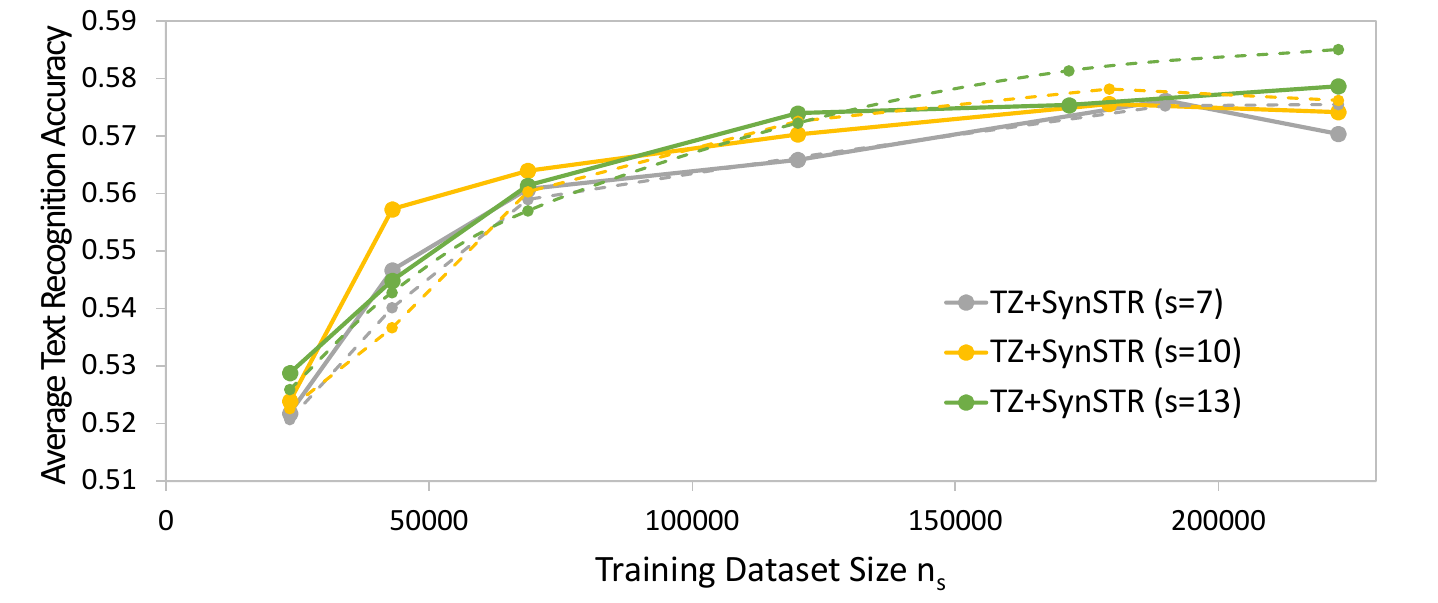}
        \subcaption{}
        \label{fig:acc_max_len_syns}
      \end{minipage}
    \end{tabular}
    \caption{Average recognition accuracy of TATT trained on the augmented datasets with various sizes. The augmented datasets were created with a different maximum word length $s\in\{7,10,13\}$. (a) and (b) show the average recognition accuracy of TATT trained on TZ+SynTZ and TZ+SynSTR, respectively. The recognitiona accuracy was evaluated by CRNN.}
\label{fig:acc_max_len}
\end{figure*}

% TATT (ssim)
\begin{figure*}[h!]
    \begin{tabular}{c}
      \begin{minipage}[t]{0.475\linewidth}
        \centering
        \includegraphics[width=8.0cm]{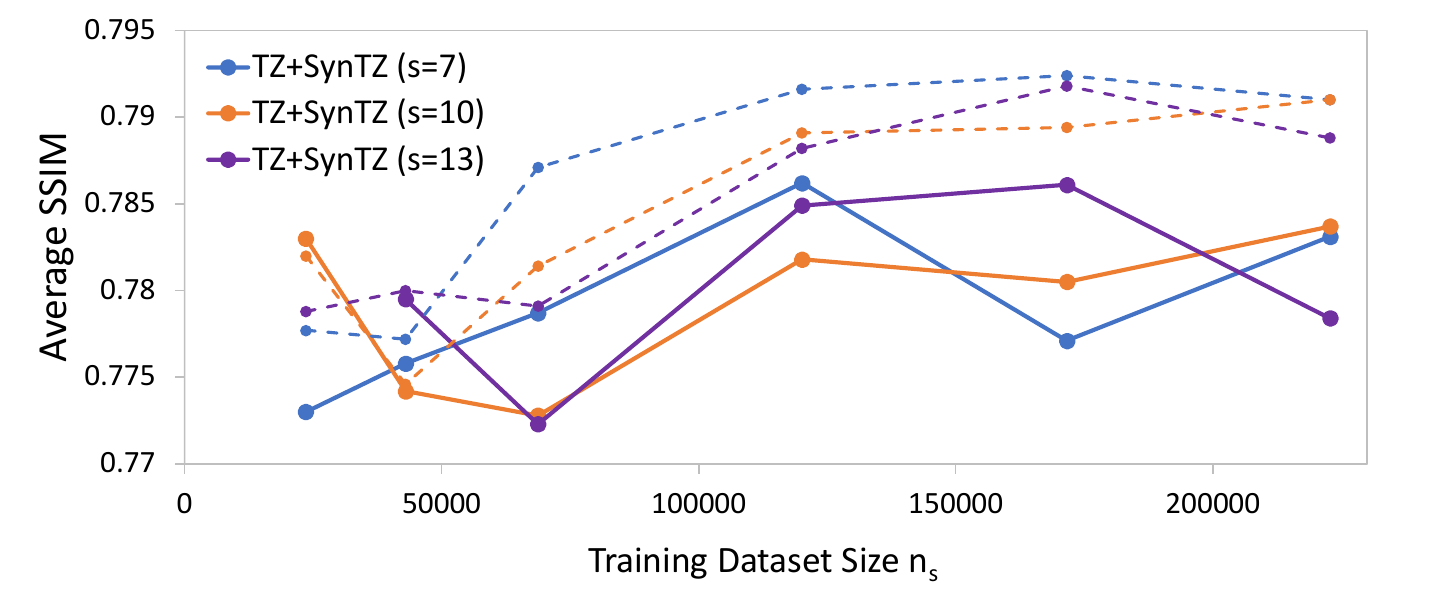}
        \subcaption{}
        \label{fig:ssim_max_len_synt}
      \end{minipage}
    
      \begin{minipage}[t]{0.475\linewidth}
        \centering
        \includegraphics[width=8.0cm]{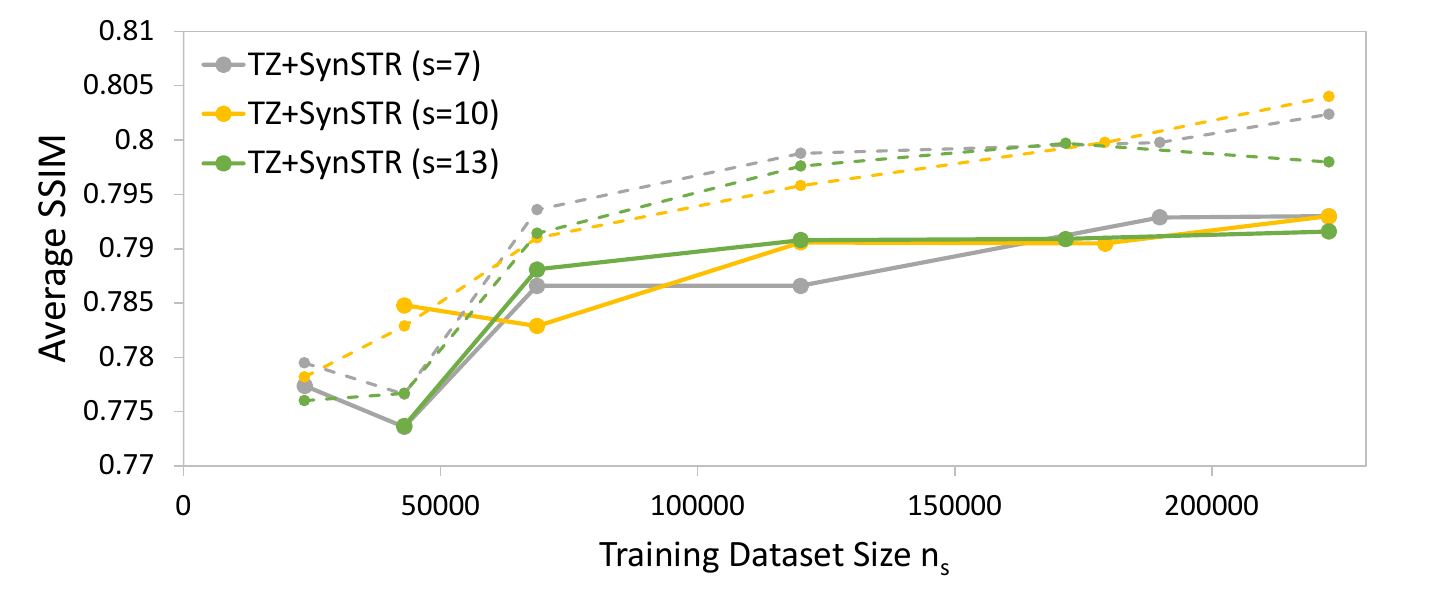}
        \subcaption{}
        \label{fig:ssim_max_len_syns}
      \end{minipage}
    \end{tabular}
    \caption{Average SSIM of TATT trained on the augmented datasets with various sizes. The augmented datasets were created with a different maximum word length $s\in\{7,10,13\}$. (a) and (b) show the average SSIM of TATT trained on TZ+SynTZ and TZ+SynSTR, respectively.}
\label{fig:ssim_max_len}
\end{figure*}

% TATT (psnr)
\begin{figure*}[h!]
    \begin{tabular}{c}
      \begin{minipage}[t]{0.475\linewidth}
        \centering
        \includegraphics[width=8.0cm]{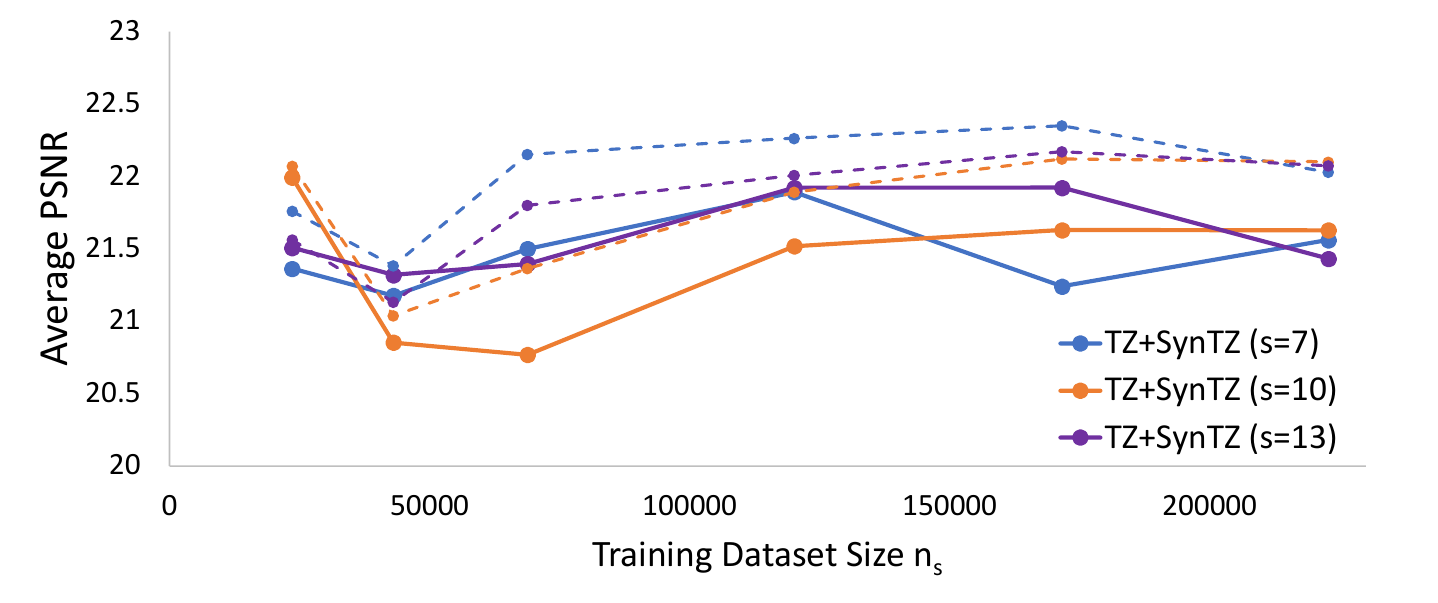}
        \subcaption{}
        \label{fig:psnr_max_len_synt}
      \end{minipage}
    
      \begin{minipage}[t]{0.475\linewidth}
        \centering
        \includegraphics[width=8.0cm]{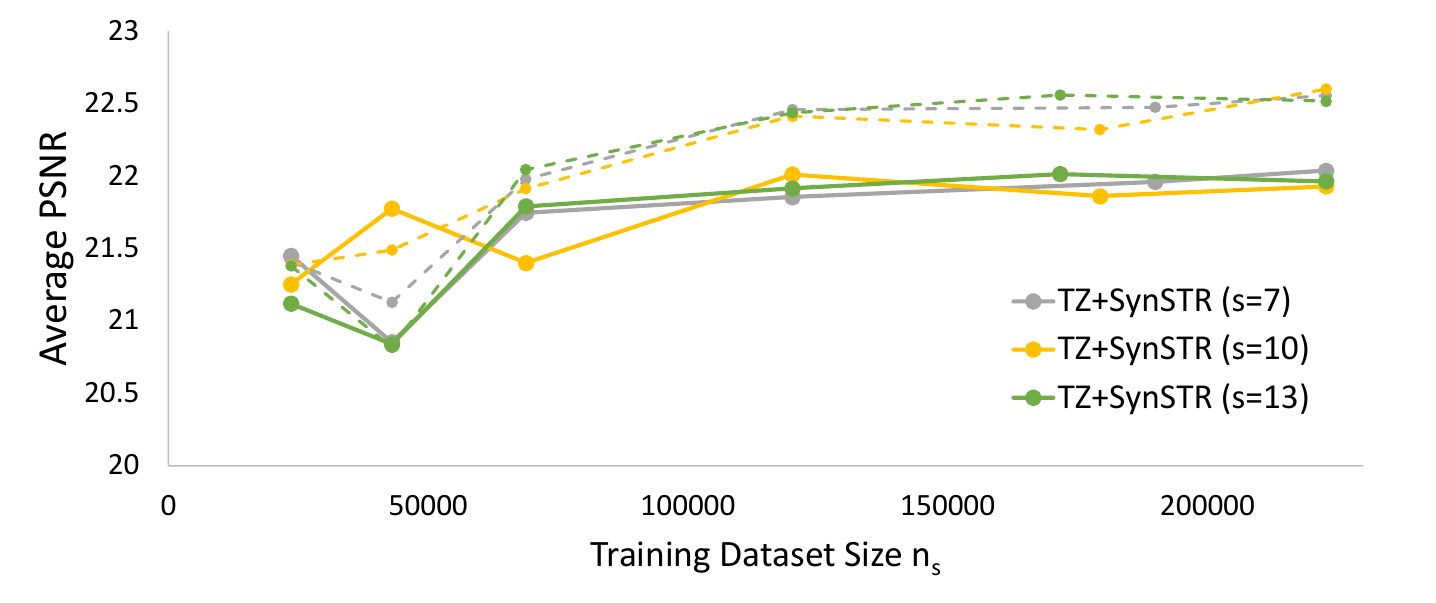}
        \subcaption{}
        \label{fig:psnr_max_len_syns}
      \end{minipage}
    \end{tabular}
    \caption{Average PSNR of TATT trained on the augmented datasets with various sizes. The augmented datasets were created with a different maximum word length $s\in\{7,10,13\}$. (a) and (b) show the average PSNR of TATT trained on TZ+SynTZ and TZ+SynSTR, respectively.}
\label{fig:psnr_max_len}
\end{figure*}

% DiMSS (acc)
\begin{figure*}[h!]
    \begin{tabular}{c}
      \begin{minipage}[t]{0.475\linewidth}
        \centering
        \includegraphics[width=8.0cm]{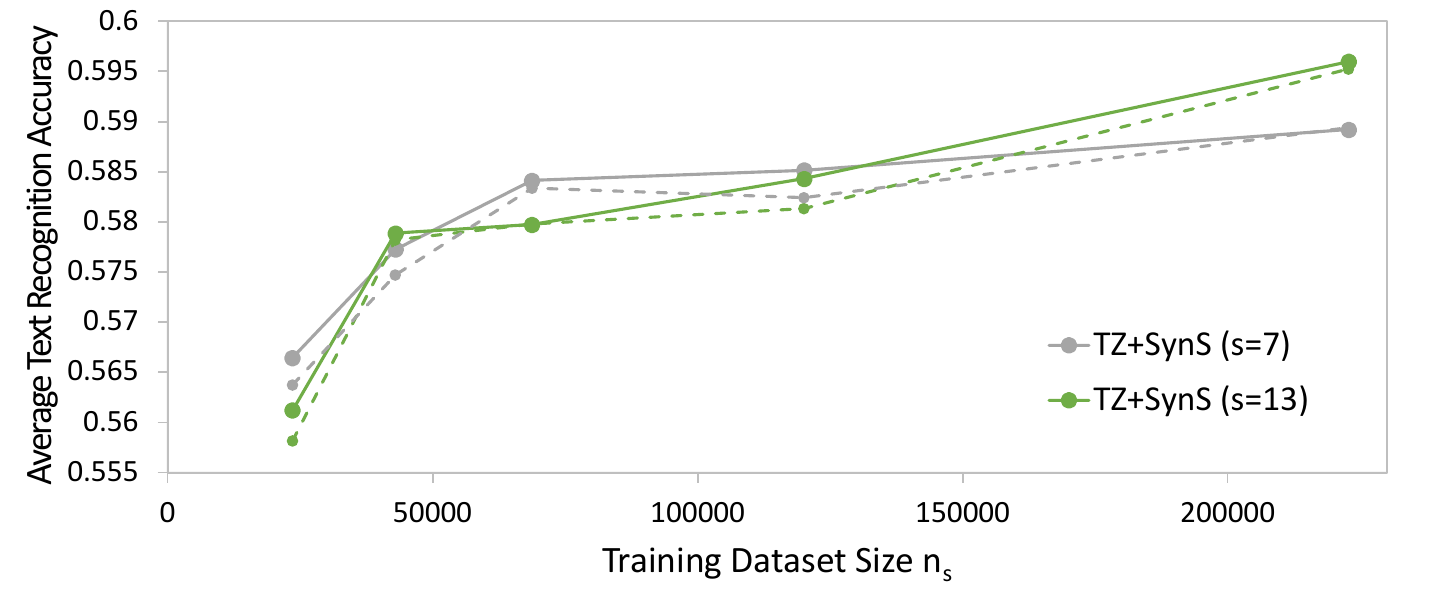}
        \subcaption{}
        \label{fig:acc_max_len_synt_dims}
      \end{minipage}
    
      \begin{minipage}[t]{0.475\linewidth}
        \centering
        \includegraphics[width=8.0cm]{figures/acc_max_len_syns_dims.pdf}
        \subcaption{}
        \label{fig:acc_max_len_syns_dims}
      \end{minipage}
    \end{tabular}
    \caption{Average recognition accuracy of our text-conditional DMs trained on the augmented datasets with various sizes. The augmented datasets were created with a different maximum word length $s\in\{7,13\}$. (a) and (b) show the average recognition accuracy of the DMs trained on TZ+SynTZ and TZ+SynSTR, respectively. The recognitiona accuracy was evaluated by CRNN.}
\label{fig:acc_max_len}
\end{figure*}

% DiMSS (ssim)
\begin{figure*}[h!]
    \begin{tabular}{c}
      \begin{minipage}[t]{0.475\linewidth}
        \centering
        \includegraphics[width=8.0cm]{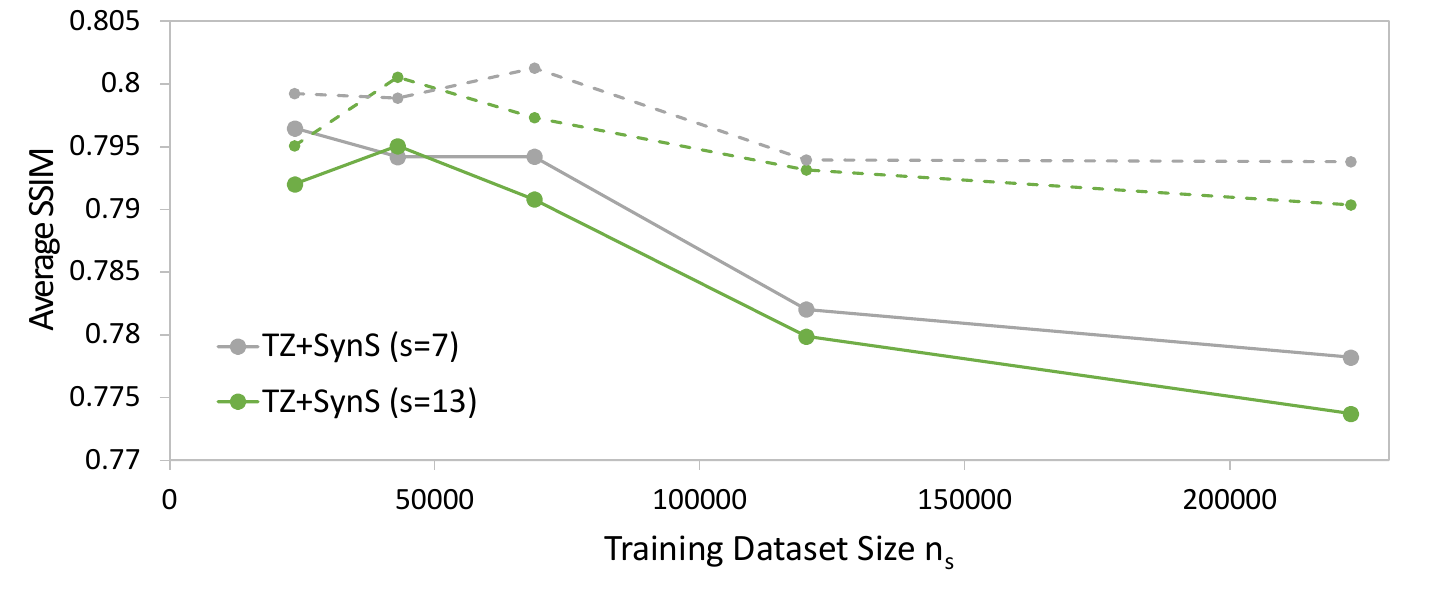}
        \subcaption{}
        \label{fig:ssim_max_len_synt_dims}
      \end{minipage}
    
      \begin{minipage}[t]{0.475\linewidth}
        \centering
        \includegraphics[width=8.0cm]{figures/ssim_max_len_syns_dims.pdf}
        \subcaption{}
        \label{fig:ssim_max_len_syns_dims}
      \end{minipage}
    \end{tabular}
    \caption{Average SSIM of our text-conditional DMs trained on the augmented datasets with various sizes. The augmented datasets were created with a different maximum word length $s\in\{7,13\}$. (a) and (b) show the average SSIM of the DMs trained on TZ+SynTZ and TZ+SynSTR, respectively.}
\label{fig:ssim_max_len}
\end{figure*}

% DiMSS (psnr)
\begin{figure*}[h!]
    \begin{tabular}{c}
      \begin{minipage}[t]{0.475\linewidth}
        \centering
        \includegraphics[width=8.0cm]{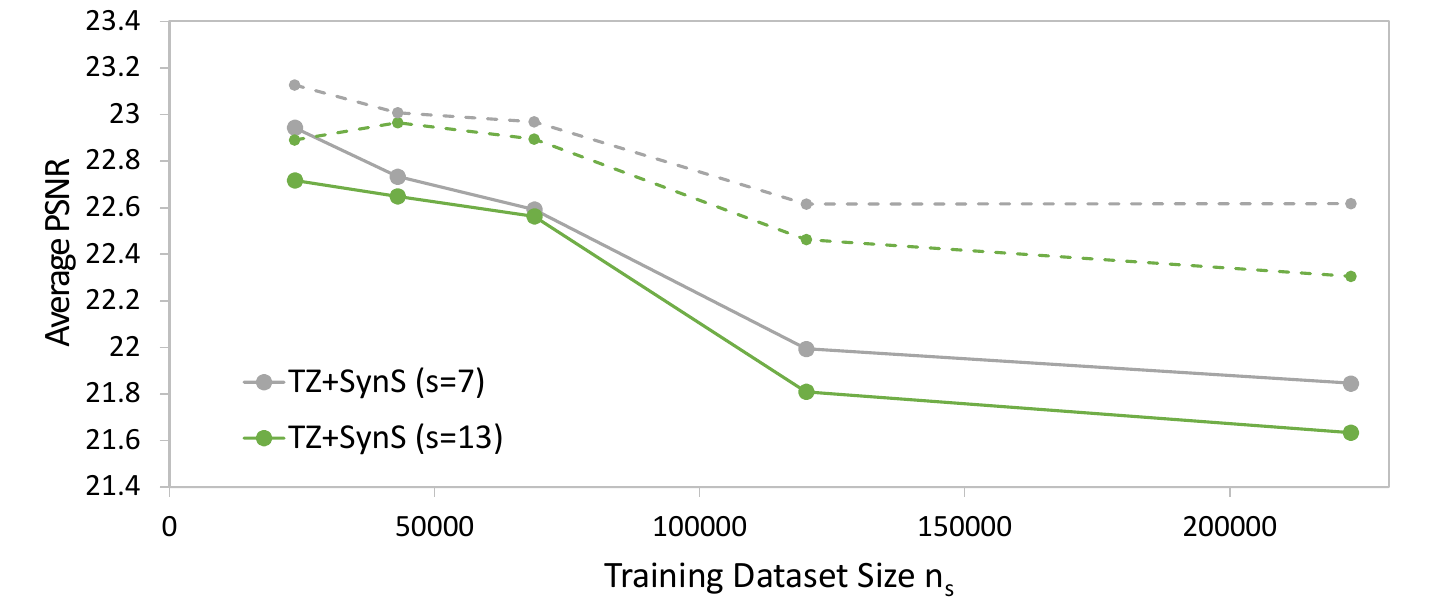}
        \subcaption{}
        \label{fig:psnr_max_len_synt_dims}
      \end{minipage}
    
      \begin{minipage}[t]{0.475\linewidth}
        \centering
        \includegraphics[width=8.0cm]{figures/psnr_max_len_syns_dims.pdf}
        \subcaption{}
        \label{fig:psnr_max_len_syns_dims}
      \end{minipage}
    \end{tabular}
    \caption{Average PSNR of our text-conditional DMs trained on the augmented datasets with various sizes. The augmented datasets were created with a different maximum word length $s\in\{7,13\}$. (a) and (b) show the average PSNR of the DMs trained on TZ+SynTZ and TZ+SynSTR, respectively.}
\label{fig:psnr_max_len}
\end{figure*}

\section{Evaluation of Extended Dataset for Varying Maximum Word Lengths}
\label{sec:word_length_evaluation}
When synthesizing text images with Synthesizer, we uniformly sampled the word length from the minimum word length of 2 to the maximum word length $s=13$ and chose a word from an English word dictionary with the same word length.
In this section, we compare several maximum word lengths $s\in\{7, 10, 13\}$ in terms of the recognition accuracy and SSIM/PSNR.
Figures \ref{fig:acc_max_len_synt}, \ref{fig:ssim_max_len_synt}, and \ref{fig:psnr_max_len_synt} show the results of recognition accuracy and SSIM/PSNR of TATT trained on TZ+SynTZ, respectively.
Figures \ref{fig:acc_max_len_syns}, \ref{fig:ssim_max_len_syns}, and \ref{fig:psnr_max_len_syns} show the same trained on TZ+SynSTR.
We can see that the recognition accuracy tends to improve as $s$ increases.
However, SSIM and PSNR did not improve depending on $s$.

In addition, Figs \ref{fig:acc_max_len_synt_dims}, \ref{fig:ssim_max_len_synt_dims}, and \ref{fig:psnr_max_len_synt_dims} show the results of recognition accuracy and SSIM/PNSR of the text-conditional DMs trained on TZ+SynTZ, respectively.
Figures \ref{fig:acc_max_len_syns_dims}, \ref{fig:ssim_max_len_syns_dims}, and \ref{fig:psnr_max_len_syns_dims} show the same trained on TZ+SynSTR.
There was no significant difference between the results for $s=7$ and $s=13$.
Therefore, following the results of TATT, we set s to 13 in the experiments in the main text.

\section{Effectiveness of Ground-truth Text Prior in Degrader}
\label{sec:effectiveness_of_texts_in_degreader}
As mentioned in the main text, ground-truth text prior was used to train Degrader.
However, determining whether the use of the ground-truth texts is effective for training Degrader is non-trivial because the image degradation process can be realized without the textual information.
Therefore, we conducted experiments to compare LR text images generated by the Degrader with and without the ground-truth text prior.
For the comparison, we trained TATT and the text-conditional DMs with HR-LR paired images of TextZoom, that is, the HR images were used as input, and the LR images were used as training targets.
To evaluate the performance, we measured SSIM/PSNR between the generated LR images and LR images of TextZoom.
Table \ref{table:effectiveness_of_ddid} shows the results of comparing TATT with TATT* and TCDM with TCDM*.
Here, TATT* and TCDM* denote TATT and the text-conditional DM trained using the ground-truth text prior, respectively.
The SSIM/PSNR of TATT* are higher than those of TATT.
In addition, the SSIM/PSNR of TCDM* are higher than those of TCDM, although by a small margin.
Based on these results, we believe that the ground-truth texts may be effective for text-image degradation.
If the variety of degradation process patterns is limited to those that appear in a specific environment, the degradation patterns of the text in the text image may be identified from the corresponding ground-truth texts to some extent.
Therefore, because our objective in this study is to imitate the TextZoom's degradation process for the dataset augmentation, we considered that the ground-truth text input would be effective for the Degrader.

\begin{table}[t!]
  \centering
  \begin{tabular}{c|cc}
    \hline
     Method & SSIM ($\times 10^{-2}$) & PSNR \\
    \hline
   TATT & 88.87  & 23.60  \\
   TATT* & \textbf{89.93} & \textbf{24.63} \\
   \hline
   TCDM & 90.12 & 25.21 \\
   TCDM* & \textbf{90.37} & \textbf{25.40}  \\
  \hline
  \end{tabular}
\caption{Comparison among different image degradation methods.}
\label{table:effectiveness_of_ddid}
\end{table}

\section{Higher-Resolution Text Images}
\label{sec:higher-resolution_text_images}

The HR text images of TextZoom contain many blurred images, which can deteriorate the performance of STISR methods.
Therefore, we can consider experiments applying Super-resolver to the HR text images of TextZoom to improve the quality of the HR images further.
Here, we refer to text images to which Super-resolver is applied as higher-resolution (HerR) text images.
First, we evaluated the HerR text images in terms of the recognition accuracy and SSIM/PSNR.
As shown in Tab. \ref{table:L_S_H_SH}, the recognition accuracy of the HerR images is further improved compared to that of the HR images.
SSIM/PSNR were measured against the corresponding HR images.

In addition, we conducted experiments in which the HerR text images were used as the target images for the TATT training, instead of the HR images .
As presented in Tab. \ref{table:tatt_hr_shr}, the recognition accuracy obtained with the HerR images is higher than that with the HR images.
On the other hand, SSIM/PSNR decreases when the HerR images were used.
The deterioration of SSIM/PSNR stems from the deviation of SSIM/PSNR between the HR and HerR images, as shown in Tab. \ref{table:L_S_H_SH}.
In practical terms, clearer images are better suited for text recognition (in fact, recognition accuracy has been improved); thus we consider that the deterioration of SSIM/PSNR due to the clearer text images is not a negative effect.

\begin{table}[t!]
  \centering
  \begin{tabular}{c|cccc}
    \hline
     Metrics & LR & SR & HR & HerR \\
    \hline
   Acc. (\%) & 26.8  & 68.1  & 72.4  & 79.3 \\
   SSIM ($\times 10^{-2}$) & 69.61 & 80.25  & -  & 86.22 \\
   PSNR & 20.35 & 22.86  & -  & 23.66 \\
  \hline
  \end{tabular}
  \caption{Comparison between LR, SR, HR, and HerR text images of TextZoom in terms of the average recognition accuracy and SSIM/PSNR. SR and HerR images were generated by applying Super-resolver to LR and HR images, respectively.}
  \label{table:L_S_H_SH}
\end{table}

\begin{table}[t!]
  \centering
    \begin{tabular}{c|cc}
        \hline
         Metrics & HR & HerR \\
        \hline
       Acc. (\%) & 52.6  & 53.81 \\
       SSIM ($\times 10^{-2}$) & 79.30 & 74.81 \\
       PSNR & 21.52 & 19.64 \\
      \hline
      \end{tabular}
\caption{Comparison of the performance of TATT trained with HR and HerR text images in terms of average recognition accuracy and SSIM/PSNR. The recognition accuracy was evaluated by CRNN.}
\label{table:tatt_hr_shr}
\end{table}

\end{document}